# INS/Odometer Land Navigation by Accurate Measurement Modeling and Multiple-Model Adaptive Estimation

Wei Ouyang, Yuanxin Wu, *Senior Member*, *IEEE,* and Hongyue Chen

*Abstract*—**Land vehicle navigation based on inertial navigation system (INS) and odometers is a classical autonomous navigation application and has been extensively studied over the past several decades. In this work, we seriously analyze the error characteristics of the odometer (OD) pulses and investigate three types of odometer measurement models in the INS/OD integrated system. Specifically, in the pulse velocity model, a preliminary Kalman filter is designed to obtain accurate vehicle velocity from the accumulated pulses; the pulse increment model is accordingly obtained by integrating the pulse velocity; a new pulse accumulation model is proposed by augmenting the travelled distance into the system state. The three types of measurements, along with the nonholonomic constraint (NHC), are implemented in the standard extended Kalman filter. In view of the motion-related pulse error characteristics, the multiple model adaptive estimation (MMAE) approach is exploited to further enhance the performance. Simulations and long-distance experiments are conducted to verify the feasibility and effectiveness of the proposed methods. It is shown that the standard pulse velocity measurement achieves the superior performance, whereas the accumulated pulse measurement is most favorable with the MMAE enhancement.**

*Index Terms*—**Land vehicle navigation, Odometer, Kalman filtering, Pulse measurement, Multiple model adaptive estimation**

## I. INTRODUCTION

THE strapdown inertial navigation system (INS) either works independently or complementarily with other sensors, e.g., the global positioning system (GPS) in case of possible signal loss and interference [1]-[13]. Under special scenarios, such as the military applications [3], the underground mining [4], and the pipeline surveying [7], the GPS signal might be not available. To fulfill autonomous navigation in GPS-denied situations, the inertial measurement units (IMU) are commonly integrated with wheel encoders/odometers (OD) to mitigate the land vehicle's navigation error drift caused by sensor biases, scale factor errors, and random walks [8], [9]. Specifically, high-precision IMUs are indispensable in order to ensure the long-distance and long-time stability of the navigation performance [13].

The INS/OD integrated system has been exhaustively investigated for decades as a typical enhancement of the strapdown inertial navigation system (SINS). Table I gives a summary of the relevant papers on the INS/OD integrated system in the last two decades. Odometers are ubiquitous in land vehicles and very convenient to be used as external measurements, but only a few studies have seriously studied the measurement model. In [1], the authors proposed to use the odometer velocity together with the vehicle motion constraints as the measurement information. However, the raw outputs of odometers are pulses that are proportional to the travelled distance and the indirectly-derived velocity outputs from the odometer pulses by approximate difference are severely corrupted by noise. To overcome this problem, the distance increments travelled over a small time period are frequently taken as measurements, which are formulated as the time integration of the velocity [8], [10]. Although the distance increment measurements are helpful to smooth out errors, more rigorous error modeling is desired to further improve the estimation accuracy. Besides, the laser doppler velocimeter (LDV), a more accurate and stable velocity sensor, has been used in lieu of the odometer for direct velocity measurement [8], [11]. Unfortunately, the LDV is vulnerable to dusty/muddy and watering roads [12]. Our team initially exploited the filtered pulse velocity as measurements [13], whereas the technical details were not disclosed and the achieved performance was unsatisfactory to us.

Researchers have also delved into studying various adaptive filtering methods to improve the INS/OD navigation performance. Leopoldo *et al.* [14] proposed an adaptive filtering technique based on the innovation sequence to account for inaccurate process and measurement covariance matrices in the localization of mobile robots. The algorithms in [15], [16], and [17] also fall into this type of the adaptive covariance matrix methods. The strong tracking Kalman filter (STF) [18] was used to keep track of the variation of the odometer scale factor error in [19] and [20]. The 'strong tracking' of parameters was realized by introducing a scaling matrix/factor into the covariance prediction process of EKF. Nevertheless, STF is

The paper was supported in part by National Key R&D Program of China (2018YFB1305103) and National Natural Science Foundation of China (61673263). A short version was presented at International Conference on Integrated Navigation Systems, Saint Petersburg, Russia, 2020.

W. Ouyang and Y. Wu are with Shanghai Key Laboratory of Navigation and Location-based Services, School of Electronic Information and Electrical Engineering, Shanghai Jiao Tong University, Shanghai 200240, China (email: ywoulife@sjtu.edu.cn, yuanx_wu@hotmail.com); H. Chen is with Beijing Institute of space launch technology, Beijing 100076, China (email: cfplyzdy@163.com ).

TABLE I
PREVIOUS RELATED RESEARCHES

| Year | Author | Technical Merits | Measurement Type* |
|------|--------|------------------|-------------------|
| 1996 | Borenstein *et al.* [24] | Measurement and correction of odometry errors. | DI |
| 2001 | Dissanayake *et al.* [1] | Odometer velocity and NHC measurements. | OV |
| 2009 | Wu *et al.* [30] | Self-calibration and observability analysis. | PV |
| 2010 | Wu *et al.* [31] | Calibration of misalignment angles. | PV |
| 2012 | Wang *et al.* [10] | Comparison of loosely and tightly coupled mode. | DI |
| 2014 | Wu [13] | Self-calibration, in-motion alignment and positioning. | PV |
| 2016 | Zhao *et al.* [19] | Adaptive two-stage Kalman filter. | OV |
| 2017 | Gao *et al.* [12] | Accurate calibration method for Laser Doppler Velocimeter. | OV |
| 2017 | Chang *et al.* [2] | Attitude estimation-based in-motion initial alignment. | OV |
| 2017 | Hidalgo-Carrió *et al.* [23] | Gaussian process estimation of odometry error. | OV |
| 2018 | Fu *et al.* [8] | Laser Doppler Velocimeter and observability analysis. | SOV |
| 2018 | Gao *et al.* [3] | Single-Axis Rotational INS/OD integrated navigation system. | OV |
| 2019 | Brossard *et al.* [22] | Learning wheel odometry and IMU errors. | OV |
| 2019 | Chen *et al.* [7] | Pipeline Surveying System and experiment tests. | OV |
| 2019 | Wang *et al.* [6] | Precise positioning of shearer based on closing path. | DI |
| 2019 | Wang *et al.* [9] | State transformation method and SINS/OD integration. | OV |

\* DI: Distance increment measurement is the output of the odometer. OV: Odometer velocity is computed by differencing the distance increment w.r.t. the sampling interval. PV: Pulse-derived velocity by using the Kalman filter. SOV: Summed odometer velocity.

only sensitive to significant change of model parameters, which is inappropriate for the INS/OD system that comprises minutely-changing parameters such as the odometer scale factor. As pointed out in [21], STF fails to detect the maneuver when the magnitude of the impulsive maneuver is small. Recently, deep learning techniques and Gaussian processes have also been applied to learn the inertial sensor error statistics or to predict the measurement residuals of the wheel odometry [22], [23]. In general, the biases of high-precision gyroscopes and accelerometers, and the odometer scale factor can be regarded as noised constants. However, the error characteristics of the odometer measurements are unknown and largely unstable [24], predictably related to vehicle motions. In this regard, the multiple model adaptive estimation technique (MMAE) [26], [25] is potentially promising to address this problem by running a bank of filters that respectively take different statistical parameters of the measurement error. The MMAE approach has been successfully applied to deal with maneuvering target tracking [27], fault detection [28] and Mars entry navigation [29], etc, in which varying parameters were involved.

Given the aforementioned problems, this article starts by analyzing the error characteristics of the odometer pulses and then three kinds of odometer measurement models are investigated. The travelled distance up to the odometer scale factor is augmented into the system state so that the accumulated pulses can be used as measurements. Furthermore, similar to the traditional distance increment method [8], [10], a counterpart is developed using the incremental pulses as measurements. The third measurement model is developed based on the pulse-derived velocity, for which a preliminary Kalman filter is formulated by assuming a constant acceleration of the vehicle forward motion in a short time interval [30], [31]. Finally, the multiple model adaptive estimation (MMAE) approach is applied to further ameliorate the performances of these methods, by exploiting the analyzed pulse error characteristics in the MMAE design process. Simulations and long-distance land vehicle experiments are conducted to validate the effectiveness of the proposed methods. Comparing with previous works, the main contributions of this article include:

1) The error characteristics of odometer pulse measurements are studied, which are conducive to the Kalman filter and the MMAE model design.
2) A new system scheme is proposed by augmenting the traditional state model with the travelled distance. In doing so, the accumulated odometer pulses could be directly used as the measurement.
3) A linear time-invariant system is used to model the pulse change over short intervals and a preliminary Kalman filter is exploited to acquire accurate velocity information. The pulse-derived velocity is shown to be quite effective in improving the navigation performance.
4) Multiple model adaptive estimation algorithms are used to even further improve the performance by accounting for the motion-dependent error characteristics of the odometer pulses.

The remaining content is organized as follows. Section II gives some preliminaries and backgrounds of INS/OD integrated navigation. Section III develops three types of measurement models and the corresponding error characteristics are investigated. Subsequently, the MMAE method is introduced in Section IV. Simulations are conducted in Section V and the results of land vehicle experiments are given in Section VI. Finally, Section VII concludes this article.

## II. PRELIMINARIES AND BACKGROUND

This section provides an overview of the INS/OD integrated navigation system. The land vehicle is equipped with a navigation-grade IMU, and the odometer is mounted on the non-steering wheel. As shown in Fig. 1, the center of the vehicle frame $O_m$ is situated at the middle point of the rear non-steering axle of the vehicle. The $x_m$ axis points forward, $y_m$ axis points upward, and $z_m$ axis is along the right direction. The odometer measures the forward motion in terms of accumulated pulses, i.e., the number of pulses generated from the very start of the vehicle motion, and we assume that the measurement frame is coincided with the vehicle frame for simplicity. The IMU frame is misaligned with the vehicle frame by mounting angles $\varphi$, $\psi$, $\theta$. The displacement between the IMU center $O_b$ and the vehicle center $O_m$ is the lever arm $\mathbf{l}^b$, which is expressed in the body frame. The navigation frame is defined as north, up and east.

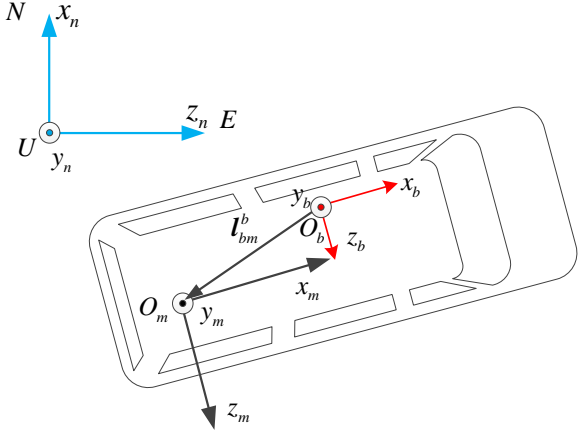

Figure 1. IMU installed on a vehicle, with definitions of the body frame, the vehicle frame, and the navigation frame.

In current INS/OD integrated navigation system, the dynamic model that includes the rate equations of the attitude, velocity and position are given as blow [13].

$$\dot{\mathbf{C}}_b^n = \mathbf{C}_b^n \left( \boldsymbol{\omega}_{nb}^b \times \right), \quad \boldsymbol{\omega}_{nb}^b = \boldsymbol{\omega}_{ib}^b - \mathbf{b}_g - \mathbf{C}_n^b \boldsymbol{\omega}_{in}^n \quad (1)$$

$$\dot{\mathbf{v}}^n = \mathbf{C}_b^n \left( \mathbf{f}^b - \mathbf{b}_a \right) - \left( 2\boldsymbol{\omega}_{ie}^n + \boldsymbol{\omega}_{en}^n \right) \times \mathbf{v}^n + \mathbf{g}^n \quad (2)$$

$$\dot{\mathbf{p}} = \mathbf{R}_c \mathbf{v}^n \quad (3)$$

where $\mathbf{C}_b^n$ is attitude matrix from the body frame to the navigation frame. $\mathbf{v}^n$ is the velocity relative to the earth. $\boldsymbol{\omega}_{ib}^b$ and $\mathbf{f}^b$ denote the error-contaminated body rate measured by the gyroscopes and the specific force measured by the accelerometers, respectively. $\boldsymbol{\omega}_{ie}^n$ is the earth rotation rate relative to the inertial frame, and $\boldsymbol{\omega}_{en}^n$ is the rotation rate of the navigation frame relative to the earth frame. $\mathbf{g}^n$ is the gravitational vector.

The vehicle's position $\mathbf{p} = \begin{bmatrix} \lambda & L & h \end{bmatrix}^T$ includes the longitude $\lambda$, latitude $L$, and height $h$. Thus, the local curvature matrix in (3) is

$$\mathbf{R}_c = \begin{bmatrix} 0 & 0 & 1/\left( R_E + h \right) \cos L \\ 1/R_N + h & 0 & 0 \\ 0 & 1 & 0 \end{bmatrix} \quad (4)$$

where $R_E$ denotes the transverse radius of curvature and $R_N$

denotes the meridian radius of curvature of the reference ellipsoid.

Besides, the odometer scale factor $K$ is defined as the number of pulses generated when the 1-m distance is travelled. The mounting angle, the lever arm, and the scale factor $K$ are regarded as random constants. For a navigation-grade IMU, the biases $\mathbf{b}_g$, $\mathbf{b}_a$ for gyroscopes and accelerometers can be modeled as constants as well.

The transformation matrix from the IMU body frame $b$ to the vehicle frame $m$, by the 2-3-1 rotation sequence, is given as

$$\mathbf{C}_b^m = \mathbf{M}_1 \left( \varphi \right) \mathbf{M}_3 \left( \theta \right) \mathbf{M}_2 \left( \psi \right) \quad (5)$$

where $\mathbf{M}_i \left( \cdot \right)$ denotes the elementary rotation matrix along the $i$-th axis. According to the observability analysis [30], the mounting angle along the forward direction is unobservable and only the mounting angles along the yaw and the pitch directions can be estimated.

Therefore, the parameters involves in the navigation system can be modeled as

$$\dot{\psi} = 0 \quad (6)$$

$$\dot{\theta} = 0 \quad (7)$$

$$\dot{K} = 0 \quad (8)$$

$$\dot{\mathbf{l}}^b = \mathbf{0} \quad (9)$$

$$\dot{\mathbf{b}}_g = \mathbf{0} \quad (10)$$

$$\dot{\mathbf{b}}_a = \mathbf{0}$$

The indirect Kalman filter is used to estimate the system error states [32]. The attitude error is defined as

$$\tilde{\mathbf{C}}_n^b = \mathbf{C}_n^b \left( \mathbf{I} + \left( \boldsymbol{\phi}^n \times \right) \right) \quad (11)$$

where $\boldsymbol{\phi}^n$ denotes the attitude error angles, and $\left( \times \right)$ represents the skew-symmetric operation.

The error format of other states are defined as the estimate subtracting the true state, i.e., $\delta \mathbf{x} = \hat{\mathbf{x}} - \mathbf{x}$. Therefore, the 21-dimension error state is

$$\mathbf{x}(t) = \begin{bmatrix} \boldsymbol{\phi}^{nT}, \delta \mathbf{v}^{nT}, \delta \mathbf{p}^T, \delta \mathbf{b}_g^T, \delta \mathbf{b}_a^T, \delta K, \delta \psi, \delta \theta, \delta \mathbf{l}^{bT} \end{bmatrix}^T \quad (12)$$

More details about the error state Kalman filter for SINS can be readily found in textbooks, e.g., [32].

## III. ODOMETER PULSE ERRORS AND MEASUREMENT MODELS

In this section, three types of measurement models are presented by means of different odometer pulse usage. Comparing with traditional forward velocity and distance increment measurements, the current models are directly based on the output of the odometer, i.e., the number of pulses. Specifically, we propose a new pulse accumulation model, in which the system state is augmented with the accumulated pulses. Similar to the traditional state-of-the-art distance increment model, the pulse increment model is then derived. At last, we report the details of the pulse-derived velocity measurement model, which have not been given in our previous work [30].

We start this section by analyzing the error characteristics of the accumulated pulses and the incremental pulses. The odometer is in nature an encoder that counts the number of pulses generated by the movement of the vehicle [34]. The odometer encoder of the land vehicle usually comprises of a detector and a number of pulse-generating physical nodes (e.g., magnetic material nodes) uniformly fixed on the rotating wheel hub. When the vehicle wheel rotates, the nodes pass in turns by the detector and a sequence of pulses are generated. Under normal scenarios without tire slips, all odometer pulses should be evenly situated across the travelled distance and the gap between two pulses is $1/K$ m, namely, one pulse corresponds to a distance increment of $1/K$ m.

As illustrated in Fig. 2(a), it is not uncommon that the detector misaligns with the pulse-generating nodes at the sampling time. However, the encoder can only count the number of pulses in integers, therefore, we define the pulse round-off error as the distance between the detector's location and the last pulse-generating node. Upon initiation, the round-off error is denoted by $\delta p_0$ (see Fig. 2(a)). Predictably, at each sampling time the detector's location relative to the pulse-generating nodes depends on the vehicle motions. In other words, the odometer pulse round-off error $\delta p_k$ is related to the vehicle motions.

Assume $N_k$ accumulated pulses have been detected at the $k$-th sampling time $t_k$. The accumulated pulse error $e_k$ can be defined as the accumulated pulse counts subtracting the true pulse counts

$$e_k = N_k - \int_0^{t_k} \dot{s}\, dt = N_{k-1} + \Delta N_k - \left( \int_0^{t_{k-1}} \dot{s}\, dt + \int_{t_{k-1}}^{t_k} \dot{s}\, dt \right)$$
$$= e_{k-1} + \Delta N_k - \int_{t_{k-1}}^{t_k} \dot{s}\, dt \quad (13)$$

in which $\Delta N_k$ is the measured pulse increment in the sampling time interval right before $t_k$, and $\dot{s}$ denotes the vehicle's forward speed in pulse/s. Especially, if the vehicle moves forward with a constant speed, we turn to have identical pulse increments $\Delta N_k$ (depending the detector's location relative to the pulse-generating nodes at both ends of the sample interval). As a result, any two consecutive accumulated pulse errors are approximately related by a constant $\Delta N_k - \dot{s}T$, where $T = t_k - t_{k-1}$. In general, however, the statistical characteristics of the accumulated pulse errors are impossible to model without the knowledge of the vehicle motion.

In order to make the problem mathematically tractable, we make an assumption below to decouple the odometer pulse measurement error with the vehicle motion and then account for their connection by using the MMAE approach in next section.

*Assumption 1:* The pulse measurement error $\delta p_k$ ($k \geq 0$), is uniformly distributed over $[0, 1)$ with variance $1/12$. In addition, the pulse measurement error is statistically independent, i.e., $E\left\{ \delta p_k \delta p_l^T \right\} = 0, i \neq k$.

### A. Pulse Accumulation Measurement

*Lemma 1:* The accumulated pulse measurement error is subject to the uniform distribution, namely,

$e_k \sim U\left( \delta p_0 - 1, \delta p_0 \right)$.

*Proof:* Assume $N_k$ accumulated pulses have been detected at the $k$-th sampling time $t_k$. As shown in Fig. 2(b), if the sampling time $t_k$ approaches the next pulse-generating time $T_{N_{k+1}}$ from left, i.e., $t_k \to T_{N_{k+1}}^-$, the true pulse counts corresponding to the travelled distance are $N_k + 1 - \delta p_0$. Therefore, the accumulated pulse measurement error is

$$e_k \to N_k - (N_k + 1 - \delta p_0) = \delta p_0 - 1 \quad (14)$$

Similarly, as $t_k \to T_{N_k}^+$, the true pulse counts corresponding to the travelled distance are $N_k - \delta p_0$. The accumulated pulse measurement error becomes

$$e_k \to N_k - (N_k - \delta p_0) = \delta p_0 \quad (15)$$

The probability that the sampling time $t_k$ happens between two pulse-generating nodes is uniform. Hence, the measurement error of the accumulated pulses is $e_k \sim U\left( \delta p_0 - 1, \delta p_0 \right)$. Besides, the variance is correspondingly computed as

$$\mathrm{var}\left(e_k\right) = \left[ \delta p_0 - (\delta p_0 - 1) \right]^2 / 12 = 1/12 \quad (16)$$

∎

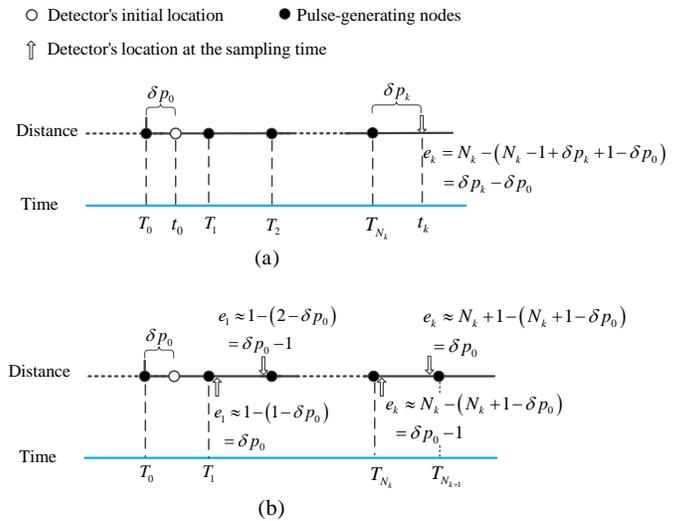

Figure 2. Definitions of round-off error and accumulated pulse measurement error on the distance line. (a) Pulse round-off error, defined as the distance between the odometer detector's location and the last pulse-generating node. (b) Upper bound and lower bound of accumulated pulse measurement error.

The time rate of travelled distance (in terms of odometer pulse $s$) is related to the system states as [30]

$$\dot{s} = K \mathbf{e}_1^T \mathbf{C}_b^m \left( \mathbf{C}_n^b \mathbf{v}^n + \boldsymbol{\omega}_{eb}^b \times \mathbf{1}^b \right) \quad (17)$$

where $\boldsymbol{\omega}_{eb}^b$ is the angular velocity of the vehicle body frame w.r.t. the earth frame. $\mathbf{e}_1$ is the 3 dimensional unit vector with the $i$-th element being 1.

In order to use the accumulated pulses as the measurement directly, the error of the travelled distance $s$ is augmented to the system error state

$$\mathbf{x}_a = \begin{bmatrix} \mathbf{x}^T & \delta s \end{bmatrix}^T \quad (18)$$

In the error-state Kalman filter, the kinematics of the pulse error should be derived and added into the dynamical model of the integrated navigation system. Specifically, we need to consider (17) with error terms, which is given as

$$\dot{\tilde{s}} = \tilde{K}\mathbf{e}_1^T \tilde{\mathbf{C}}_b^m \left( \tilde{\mathbf{C}}_n^b \tilde{\mathbf{v}}^n + \tilde{\boldsymbol{\omega}}_{eb}^b \times \tilde{\mathbf{l}}^b \right) \tag{19}$$

$$\tilde{K} = K + \delta K \tag{20}$$

$$\tilde{\mathbf{C}}_b^m = \left( \mathbf{I} - \delta \boldsymbol{\alpha} \times \right) \mathbf{C}_b^m \tag{21}$$

$$\tilde{\mathbf{l}}^b = \mathbf{l}^b + \delta \mathbf{l}^b \tag{22}$$

where the IMU-vehicle misalignment angles is denoted by $\delta \boldsymbol{\alpha} = \left[ \delta \varphi, \delta \psi, \delta \theta \right]^T$.

In (19), $\tilde{\boldsymbol{\omega}}_{eb}^b$ can be further expanded as

$$\tilde{\boldsymbol{\omega}}_{eb}^b = \tilde{\boldsymbol{\omega}}_{ib}^b - \tilde{\mathbf{C}}_n^b \tilde{\boldsymbol{\omega}}_{ie}^n \tag{23}$$

$$\tilde{\boldsymbol{\omega}}_{ib}^b = \boldsymbol{\omega}_{ib}^b + \mathbf{b}_g \tag{24}$$

$$\tilde{\mathbf{C}}_n^b \tilde{\boldsymbol{\omega}}_{ie}^n = \mathbf{C}_n^b \left( \mathbf{I} + \left( \boldsymbol{\phi}^n \times \right) \right) \left( \boldsymbol{\omega}_{ie}^n + \delta \boldsymbol{\omega}_{ie}^n \right) \tag{25}$$

where $\delta \boldsymbol{\omega}_{ie}^n$ reflects the influence of the position error.

$$\delta \boldsymbol{\omega}_{ie}^n = \frac{\partial \boldsymbol{\omega}_{ie}^n}{\partial \mathbf{p}} \delta \mathbf{p} = \omega_{ie} \begin{bmatrix} 0 & -\sin L & 0 \\ 0 & \cos L & 0 \\ 0 & 0 & 0 \end{bmatrix} \delta \mathbf{p} \tag{26}$$

in which $\omega_{ie}$ is the rotational rate of the Earth.

Substitute (20)-(26) into (19), the kinematics of the pulse error is obtained

$$\delta \dot{s} = \dot{\tilde{s}} - \dot{s} \tag{27}$$

The Jacobian matrices are obtained by calculating the partial derivative of (27) w.r.t. error states, which are given in Appendix A.

Accompanied with the NHC constraint [13], the complete measurement model is

$$\begin{bmatrix} y_s \\ \mathbf{y}_{nhc} \end{bmatrix} = \begin{bmatrix} \mathbf{e}_2^T \\ \mathbf{e}_3^T \end{bmatrix} \mathbf{C}_b^m \left( \mathbf{C}_n^b \mathbf{v}^n + \boldsymbol{\omega}_{eb}^b \times \mathbf{l}^b \right) \end{bmatrix} \tag{28}$$

The corresponding measurement matrix is then computed as

$$\mathbf{H}_s = \begin{bmatrix} \begin{bmatrix} \mathbf{0}_{1 \times 21} & 1 \end{bmatrix} \\ \mathbf{H}_{nhc} \end{bmatrix} \tag{29}$$

Note that the Jacobian matrices w.r.t. the NHC are similarly computed as the method given in Appendix A. The difference is that $\mathbf{e}_1$ should be replaced with $\mathbf{e}_2$ and $\mathbf{e}_3$, respectively.

The measurement prediction of accumulated number of pulse counts requires the integration of (17)

$$s_{k+1} = s_k + K\mathbf{e}_1^T \mathbf{C}_b^m \int_{t_k}^{t_{k+1}} \mathbf{C}_n^b(t) \mathbf{v}^n(t) + \boldsymbol{\omega}_{eb}^b(t) \times \mathbf{l}^b dt \tag{30}$$

The integral involved can be expanded using (23)

$$\int_{t_k}^{t_{k+1}} \mathbf{C}_n^b(t) \mathbf{v}^n(t) + \boldsymbol{\omega}_{eb}^b(t) \times \mathbf{l}^b dt$$
$$= \int_{t_k}^{t_{k+1}} \mathbf{C}_n^b(t) \mathbf{v}^n(t) + \boldsymbol{\omega}_{ib}^b(t) \times \mathbf{l}^b - \left( \mathbf{C}_n^b(t) \boldsymbol{\omega}_{ie}^n \right) \times \mathbf{l}^b dt \tag{31}$$

An accurate calculation of (31) is provided in Appendix B.

## B. Pulse Increment Measurement

The second type of measurement is the pulse increment in time intervals of interest. The dynamical model is the same as in Section II. The pulse increment measurement model is

$$\Delta s = \sum_{k=1}^{N} \int_{t_k}^{t_{k+1}} \dot{s} \, dt$$
$$= \sum_{k=1}^{N} K\mathbf{e}_1^T \mathbf{C}_b^m \int_{t_k}^{t_{k+1}} \mathbf{C}_n^b(t) \mathbf{v}^n(t) + \boldsymbol{\omega}_{eb}^b(t) \times \mathbf{l}^b dt \tag{32}$$

Note that the pulse increment is related to all states during the considered interval, instead of the state at a fixed time. This violates the basic assumption of the standard Kalman filter that the measurement is simply a function of the current state.

Along with the NHC measurements, the complete measurement model is given as

$$\begin{bmatrix} y_{\Delta s} \\ \mathbf{y}_{nhc} \end{bmatrix} = \begin{bmatrix} \Delta s \\ \begin{bmatrix} \mathbf{e}_2^T \\ \mathbf{e}_3^T \end{bmatrix} \mathbf{C}_b^m \left( \mathbf{C}_n^b \mathbf{v}^n + \boldsymbol{\omega}_{eb}^b \times \mathbf{l}^b \right) \end{bmatrix} \tag{33}$$

Here, we need to derive the measurement matrix of the pulse increment measurement. Similar to the method used in Appendix A, the error format for (32) is

$$\delta \Delta s \approx \sum_{k=1}^{N} \int_{t_k}^{t_{k+1}} \mathbf{M}_k \delta \mathbf{x}_k dt$$
$$\approx \sum_{k=1}^{N} \int_{t_k}^{t_{k+1}} \mathbf{M}_k \boldsymbol{\Phi}_{k|N} \delta \mathbf{x}_N dt$$
$$= \left( \sum_{k=1}^{N} \int_{t_k}^{t_{k+1}} \mathbf{M}_k dt \boldsymbol{\Phi}_{k|0} \right) \boldsymbol{\Phi}_{0|N} \delta \mathbf{x}_N \tag{34}$$

where $\mathbf{M}_k$ denotes the Jacobian matrix of pulse velocity w.r.t. state $\mathbf{x}$ at $t_k$. $\boldsymbol{\Phi}_{k|k-1}$ denotes the state transition matrix from $t_{k-1}$ to $t_k$ that is assumed to be constant during the small time interval $T$. The integration of $\mathbf{M}_k$ in (34) is denoted as $\mathbf{H}_k$, and its derivation is provided in Appendix C. Therefore, the complete measurement matrix for pulse increment measurement is

$$\mathbf{H}_{\Delta s} = \begin{bmatrix} \left( \sum_{k=1}^{N} \mathbf{H}_k \boldsymbol{\Phi}_{k|0} \right) \boldsymbol{\Phi}_{0|N} \\ \mathbf{H}_{nhc} \end{bmatrix} \tag{35}$$

If the vehicle is driven mildly, $\mathbf{M}_k$ can be regarded as constant over the small interval of length $T$. In addition, if no fast turns exist in the interval of length $NT$, the state transition matrices can be further approximated by an identity matrix. In this case, (35) can be further approximated as

$$\mathbf{H}_{\Delta s} = \begin{bmatrix} \sum_{k=1}^{N} \mathbf{M}_k T \\ \mathbf{H}_{nhc} \end{bmatrix} \tag{36}$$

Additionally, the measurement prediction of (32) is computed following the Appendix B.

*Lemma 2:* The pulse increment measurement error $i_k$ satisfies $E\left\{ i_k i_{k-1}^T \right\} = -1/12$ and $\mathrm{var}\left( i_k \right) = 1/6$.

*Proof:* The accumulated pulse measurement error is

computed as

$$i_k = \left( \tilde{s}_k - \tilde{s}_{k-1} \right) - \left( s_k - s_{k-1} \right) = \delta p_k - \delta p_{k-1} \quad (37)$$

Since $\delta p_k \sim U\left(0,1\right)$, according to Assumption 1 and (16),

$$E\left\{ i_k i_{k-1}^T \right\} = E\left\{ \left( \delta p_k - \delta p_{k-1} \right)\left( \delta p_{k-1} - \delta p_{k-2} \right)^T \right\} \\ = -1/12 \quad (38)$$

which means that the increment measurement errors are not independent.

In addition, the variance of the incremental measurement error is computed as

$$\mathrm{var}\left( i_k \right) = E\left\{ i_k i_k^T \right\} = E\left\{ \left( \delta p_k - \delta p_{k-1} \right)\left( \delta p_k - \delta p_{k-1} \right)^T \right\} \\ = 1/6 \quad (39)$$

∎

### C. Pulse Velocity Measurement

The pulse velocity information is usually calculated from the pulse increment over the sampling interval by

$$\dot{\tilde{s}}_k = \frac{\Delta N_k + i_k}{T} \quad (40)$$

where $\Delta N_k$ is the measured incremental pulse counts and $i_k$ is the corresponding pulse increment measurement error.

If the wheel slipping and skidding are not considered, the pulse increment error satisfies the property given in Lemma 2. Unfortunately, even in normal situations (40) will still introduce severe noise into the velocity as the sampling interval $T$ is quite short. For instance, if $i_k = 0.5$ and $T = 0.02$s, then the velocity error is 25 pulse/s, which is too inaccurate to be used.

To circumvent this problem, a preliminary Kalman filter is adopted to derive the velocity by using the accumulated pulses as observations. Specifically, we assume a constant acceleration rate of the vehicle forward motion, namely,

$$\begin{bmatrix} ds/dt \\ d\dot{s}/dt \\ d\ddot{s}/dt \end{bmatrix} = \begin{bmatrix} 0 & 1 & 0 \\ 0 & 0 & 1 \\ 0 & 0 & 0 \end{bmatrix}\begin{bmatrix} s \\ \dot{s} \\ \ddot{s} \end{bmatrix} + \begin{bmatrix} 0 \\ 0 \\ 1 \end{bmatrix} w \quad (41)$$

$$y_k = s_k + e_k \quad (42)$$

where $w$ is the dynamic model error and $e_k$ is the accumulated pulse measurement error. Note that the error characteristics of the estimated pulse velocity is influenced by the vehicle motions that have not been modeled in (41).

Upon obtaining the pulse velocity information, the complete measurement model is constructed as

$$\begin{bmatrix} \dot{s} \\ \mathbf{y}_{nhc} \end{bmatrix} = diag\left( \begin{bmatrix} K & 1 & 1 \end{bmatrix} \right) \mathbf{C}_b^m \left( \mathbf{C}_n^b \mathbf{v}^n + \boldsymbol{\omega}_{eb}^b \times \mathbf{l}^b \right) \quad (43)$$

The corresponding measurement matrix is

$$\mathbf{H}_{\dot{s}} = \begin{bmatrix} \mathbf{M}_k \\ \mathbf{H}_{nhc} \end{bmatrix} \quad (44)$$

where the pulse velocity measurement matrix $\mathbf{M}_k$ has been developed in (34).

## IV. Accounting for Vehicle Motion Dependence by MMAE Method

Considering that the error characteristics of three types of measurements are varying with the motion, the multiple model method is selected from a wealth of adaptive algorithms to deal with this problem. In this work, the standard deviation of the odometer measurement error is taken as the modeling parameter and the theoretical characteristics of pulse measurement errors in Section III are exploited to guide the design of the candidate models.

The rationale behind the MMAE method is the adaptive selection of model parameters according to the probability density function (pdf) [36]. Assume that the model set includes $M$ models and each model is parameterized with $\mathbf{p}^{(j)}$. Initially, each model is assigned with equal weight, and then gradually updated based on the measurement residual and residual covariance. The process of weight updating and normalizing is given as

$$w_k^{(j)} = w_{k-1}^{(j)} p\left( \tilde{\mathbf{y}}_k \mid \tilde{\mathbf{x}}_k^{-(j)} \right) \\ w_k^{(j)} \leftarrow \frac{w_k^{(j)}}{\sum\limits_{j=1}^{M} w_k^{(j)}} \quad (45)$$

in which $\leftarrow$ denotes replacement, and $M$ is the number of models.

And, the pdf of each residual is computed as

$$p\left( \tilde{\mathbf{y}}_k \mid \tilde{\mathbf{x}}_k^{-(j)} \right) = \frac{1}{\left[ \det\left( 2\pi \mathbf{S}_k^{(j)} \right) \right]^{1/2}} \exp\left\{ -\frac{1}{2} \mathbf{e}_k^{(j)T}\left( \mathbf{S}_k^{(j)} \right)^{-1} \mathbf{e}_k^{(j)} \right\} \quad (46)$$

where $\mathbf{e}_k^{(j)}$ and $\mathbf{S}_k^{(j)}$ are respectively the residual and the corresponding covariance matrix of the $j$-th model.

The state and covariance matrix estimated from the MMAE method are computed by

$$\hat{\mathbf{x}}_k^+ = \sum_{j=1}^{M} w_k^{(j)} \hat{\mathbf{x}}_k^{+(j)} \\ \mathbf{P}_k^+ = \sum_{j=1}^{M} w_k^{(j)} \left[ \left( \hat{\mathbf{x}}_k^{+(j)} - \hat{\mathbf{x}}_k^+ \right)\left( \hat{\mathbf{x}}_k^{+(j)} - \hat{\mathbf{x}}_k^+ \right)^T + \mathbf{P}_k^{+(j)} \right] \quad (47)$$

where $\hat{\mathbf{x}}_k^{+(j)}$ and $\mathbf{P}_k^{+(j)}$ are the estimated state and covariance matrix in terms of the $j$-th model, respectively. Here, superscripts '$-$' and '$+$' denote the predicted and updated values, respectively.

The adapted model parameter is calculated by

$$\hat{\mathbf{p}}_k = \sum_{j=1}^{M} w_k^{(j)} \mathbf{p}^{(j)} \quad (48)$$

The MMAE method performs well as long as one model uses the correct or nearly correct parameters [37]. As a result, the theoretical analysis and empirical knowledge could assist the design of the model set for practitioners.

## V. Simulation Results

This section conducts simulations to verify the feasibility and effectiveness of proposed methods. The vehicle is equipped with a navigation-grade IMU, which includes a triad of gyroscopes (bias $0.005°/h$, noise $0.001°/\sqrt{h}$) and accelerometers (bias $30\mu g$, noise $5\mu g/\sqrt{Hz}$). The odometer scale factor is 59.8 p/m. The IMU mounting angles are 3 deg in the yaw direction and 2 deg in the pitch direction. The lever arm is set to $\mathbf{l}^b = [1, 0.5, 0.8]^T$ m. The vehicle goes forward with four kinds of moving patterns: constant acceleration or deceleration, turns, constant velocity, and varying acceleration or deceleration. The simulation time is 5000 seconds and the IMU update interval $T = 0.02s$. The trajectory is illustrated in Fig. 3(a). It can be seen that five turns are conducted and the route is ended with a long straight line. Periodic acceleration/deceleration are performed when the vehicle drives forward. Fig. 3 (b) shows the acceleration/deceleration history in the first two periods, i.e., 240 seconds. Specifically, in addition to the acceleration and deceleration, the vehicle moves forward with constant speeds during the intervals 0~10s, 20~80s, and 90~120s of each period. Note that a sine function is applied to simulate the varying acceleration/deceleration, and the initial pulse round-off error $\delta p_0$ is set as zero.

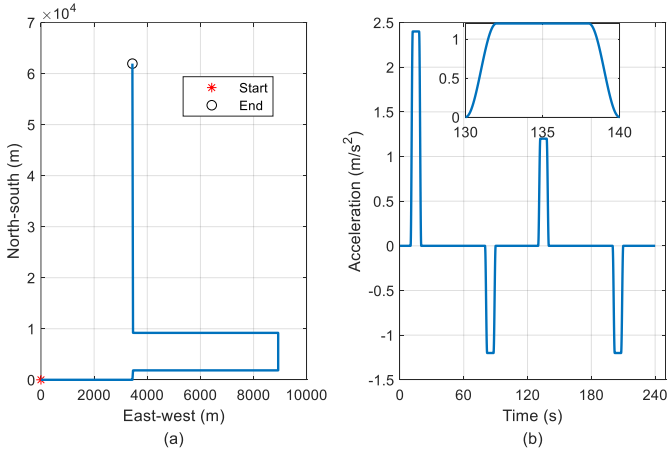

Figure 3: The vehicle trajectory and acceleration history. (a) A trajectory of about 76 km with five turns. (b) Acceleration and deceleration profile in the first 240 seconds. The subfigure shows the sine-form acceleration/deceleration in detail.

We first examine the accumulated pulse error and the pulse increment error under different moving patterns. As shown in the first row of Fig. 4, the distribution of accumulated pulse errors is not strictly uniform in contrast to Lemma 1. When the vehicle moves with constant speeds (0~10s, 20~80s, 90~120s), two arbitrary consecutive accumulated pulse errors are roughly related with a constant. In the middle row of Fig. 4, it can be seen that the pulse increment error is neither normally nor uniformly distributed. With uniform speeds, the pulse increment errors are situated at −0.2 and 0.8 due to the correlation between two consecutive errors as discussed in Lemma 1 and Lemma 2. It should be stressed that in the context of practical vehicle movement the error characteristics of accumulated pulses and incremental pulses will be much more complicated than what we have observed in Fig. 4.

The accuracy of the pulse velocity obtained by the preliminary Kalman filter is also examined. The third row of Fig. 4 shows that satisfactory accuracy can be achieved other than the time when varying acceleration and deceleration are experienced. These large errors are caused by the inaccurate modeling of pulse in (41), in which the varying acceleration rate is not considered. It also indicates that the pulse velocity errors are almost surely smaller than 0.5 p/s during the time with constant acceleration rates. However, varying acceleration and deceleration are common in normal vehicle movement. Large pulse velocity errors will be encountered frequently in practice, and of course the error characteristics also depend on the vehicle motions. In addition, we also notice that the pulse velocity error is also influenced by the magnitude of the odometer scale factor. The encoder with higher resolution tends to generate larger pulse velocity errors during the time with changing acceleration and deceleration. This is because the magnitude of pulse velocity becomes larger with higher odometer resolution, and the estimation error will be accordingly enlarged when varying acceleration and deceleration are experienced.

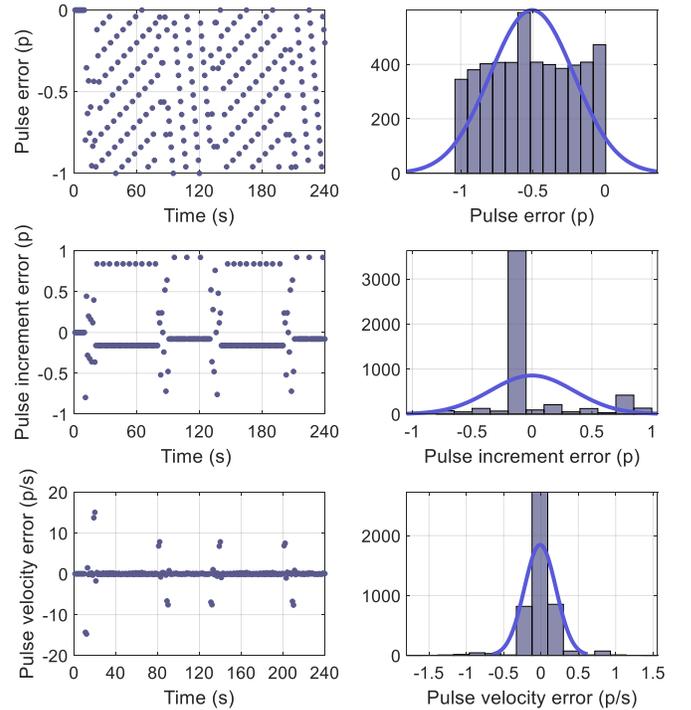

Figure 4: Error characteristics of three types of pulse measurements. Top row: accumulated pulse errors; middle row: pulse increment errors; bottom row: pulse velocity obtained from the Kalman filter. Note that the bottom right figure only shows the pulse velocity errors smaller than 2 p/s.

As shown in Fig. 5, the position estimation errors of three measurement types are about 0.01% of the travelled distance, which are satisfactory for this application and much better than the acceptable maximum error ratio 0.1%. In contrast, the pulse velocity type slightly outperforms the other two types. However, the pulse velocity type significantly relies on the detection of large measurement errors, which is fulfilled by setting a threshold on the EKF innovation. Subsequently, the standard deviation is appropriately enlarged in the Kalman filter. According to Fig. 4, the initial standard deviation is set to 0.5 and the innovation threshold is set to 1.5. Measurements out of the innovation threshold are treated with a larger standard deviation of 5. By contrast, the measurement errors of the pulse accumulation and pulse increment are relatively smaller and the standard deviation is set to 1 for both of them. Note that the standard deviation of the measurement in the navigation filter should be compatible with the resolution of the odometer. For example, the standard deviation should be set as larger values while using the odometer with higher resolutions.

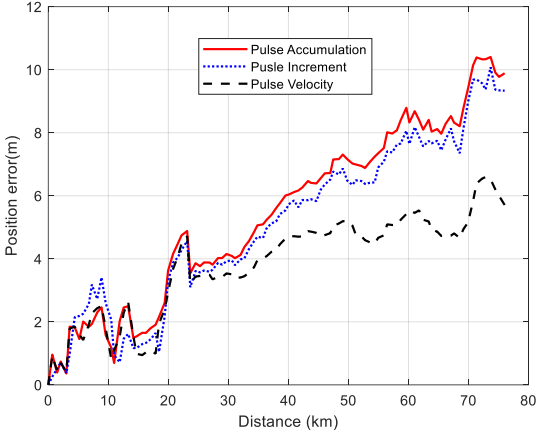

Figure 5: Horizontal position estimate errors.

The system parameter estimation results are shown in Fig. 6-7. As shown in Fig. 6, the results are similar for three measurement types; however, the odometer scale factor and mounting angles are slightly more accurate using the pulse accumulation measurements. The lever arm results in Fig. 7 reveal that the pulse velocity measurement performs the best whereas the pulse increment method is biased in the forward lever arm estimate. We found that this was caused by the negligence of the correlation in pulse increment measurement errors, which can be fixed by using the Stochastic Cloning Kalman filter (SCKF) [33]. The position estimation accuracy of SCKF is nearly equivalent to EKF, therefore, we still adopted it in the current work. The estimated IMU parameters are shown in Fig. 8. The results are similar for three measurement types, and the accelerometer bias is apparently more observable than the gyroscope bias.

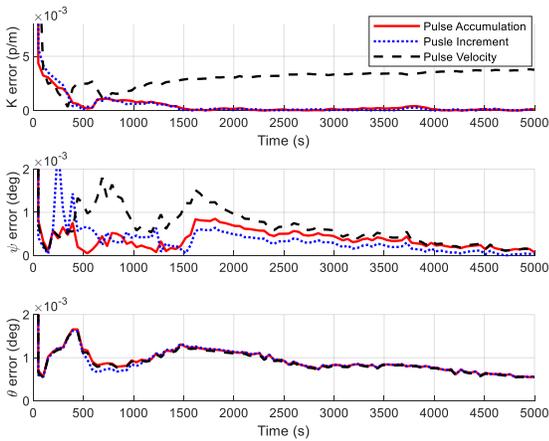

Figure 6: Odometer scale factor and mounting angle errors.

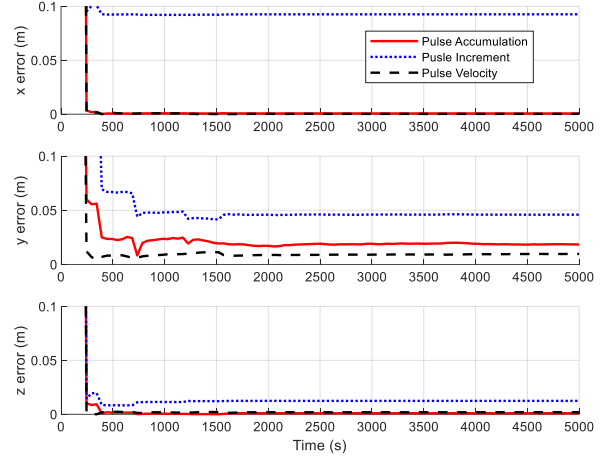

Figure 7: Lever arm errors.

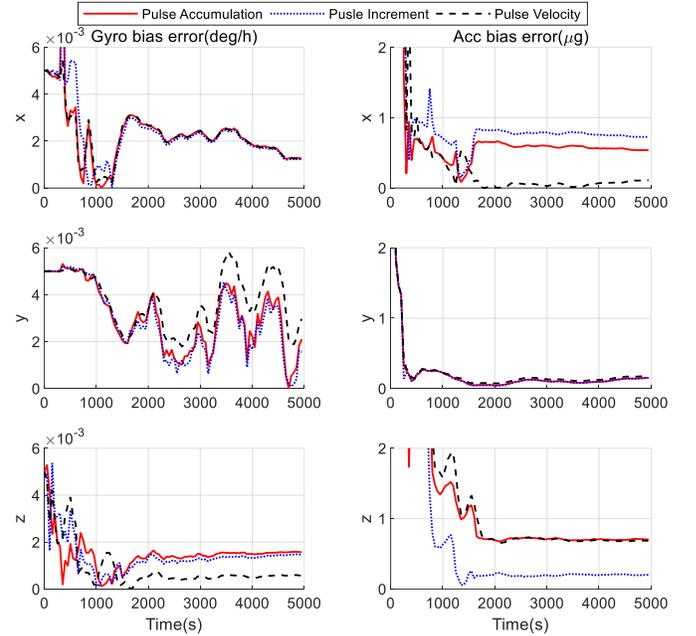

Figure 8: Gyroscope and accelerometer biases errors.

In order to highlight the effectiveness of MMAE, noises with different magnitude of standard deviations are added to the true measurements. As the theoretical analysis shows, the ideal standard deviation for pulse accumulation and increment errors is sqrt(1/12) ≈ 0.3p and sqrt(1/6) ≈ 0.41p. Therefore, the noises with standard deviations 0.5p (0~1000s), 2p (1000~3000s), and 5p (3000~5000s) are added to the accumulated and incremental pulse measurements. In contrast, the magnitude of pulse velocity errors is larger than the other two types of errors. The noises with standard deviations 0.5p/s (0~1000s), 5p/s (1000~3000s), and 20p/s (3000~5000s) are added to the pulse velocity measurements. Figs. 9-11 show the results after using the MMAE method, in which the ideal model sets are applied. However, the standard methods use the smallest standard deviation with a reasonable assumption that the measurement information is reliable. In general, the prior knowledge of the measurement errors is hard to access. It can be seen that larger measurement error will deteriorate the estimation performance in standard methods, whereas the MMAE technique can adaptively select the most appropriate standard deviation (as observed in the lower subfigures) and effectively suppress the error accumulation.

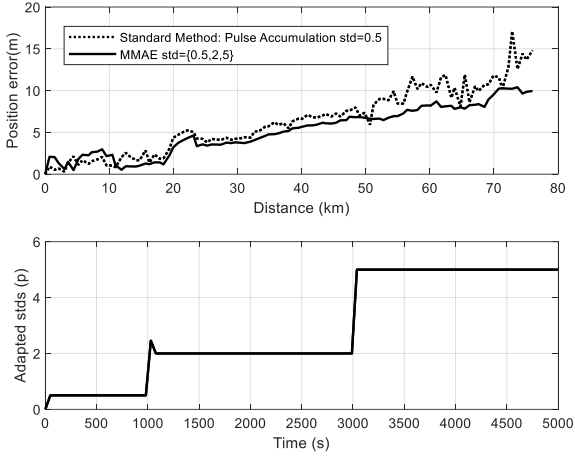

Figure 9: Pulse accumulation measurements: comparison of standard method and MMAE.

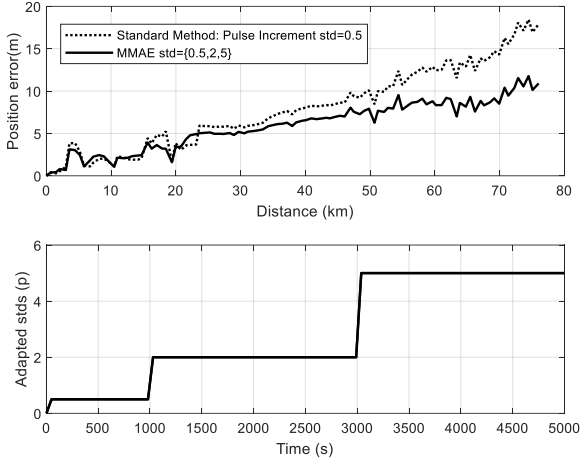

Figure 10: Pulse increment measurements: comparison of standard method and MMAE.

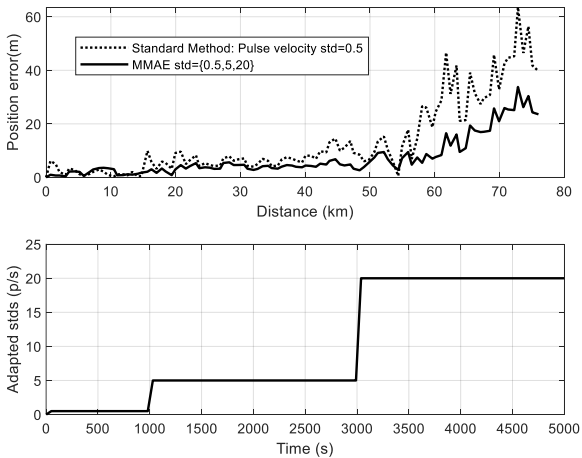

Figure 11: Pulse velocity measurements: comparison of standard method and MMAE.

## VI. FIELD TEST RESULTS

Land vehicle experiments were conducted to test the long-time and long-distance performance of proposed methods. The vehicle was equipped with a navigation-grade IMU set and an odometer with a scale factor about 53 p/m. The bias stability and random walk for gyroscopes are normally $0.01°/h$ and $0.002°/\sqrt{h}$, respectively. For accelerometers, the bias stability is $50\mu g$, and the random walk is $10\mu g/\sqrt{Hz}$. The sampling frequency of IMU is 100Hz and the 2-sample algorithm was exploited in the navigation solution. Besides, the odometer pulses generated in each $T = 0.02s$ were stored as measurements. The pulse accumulation measurements were the direct summation of all pulses from the very start. After that, the pulse velocity measurements were obtained through the preliminary Kalman filter proposed in Section III.

In order to corroborate the effectiveness of proposed methods, we conducted two runs on the same route, as shown in Fig. 12. In each experiment, the vehicle was kept still for 200~500 seconds for the initial alignment.

The reference data were obtained using the INS/OD/GPS integrated navigation. The position accuracy of GPS was about 2m (1σ) and the accuracy of the reference position was better than 0.5m (1σ). As the height error of this system is ready to be aided with an atmospheric pressure altimeter [8], a reason that the height accuracy is usually not listed for commercialized products [35], the system performance is mainly evaluated by the horizontal position accuracy.

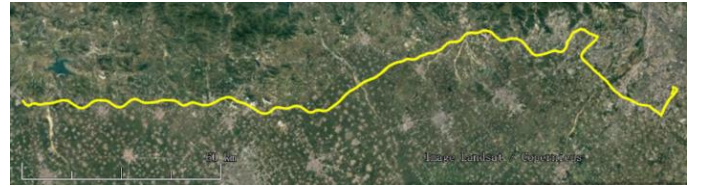

Figure 12: Field tests last about 7 hours with the distance about 490km

### A. Online Calibration and Navigation

We at first examine the methods proposed in Section III. Suppose no prior information about the system parameters is available. The odometer scale factor, lever arm, mounting angles, and IMU biases were all initialized as zeroes in the following navigation algorithms. The Kalman filter update interval is 1s for three types of measurements. In order to avoid the adverse effect of abnormal measurements caused by possible wheel slipping and skidding, we routinely conducted the variance relaxation when abnormal innovations were detected. More details about using three types of measurements are given as follows:

#### 1) Pulse accumulation measurement

As shown in Lemma 1, the theoretical measurement error of accumulated pulse should be uniformly distributed with variance 1/12. In simulations, small measurement variance could be used. In field tests, larger standard deviations such as 0.5 or 1 are favored. Moreover, we also found that smaller NHC variance was preferred for accumulated pulses, which puts more strict constraints on the vehicle's orientation.

## 2) *Pulse increment measurement*

In field tests, the vehicle's speed was mostly 60~100 km/h and more than one thousand incremental pulses were generated in 1s. Therefore, the standard deviation of the measurement error was set to 2 to account for other systematic errors.

## 3) *Pulse velocity measurement*

The pulse velocity could be estimated from the accumulated pulses by the preliminary Kalman filter and used as the measurement. The simulation result indicated that large pulse velocity error would be generated once varying acceleration and deceleration were experienced. As a remedy for this drawback, a threshold was set on the innovation and a larger standard deviation was used to cope with these inaccurate measurements. It was also conducive to avoid the adverse effect of abnormal measurements induced by the slipping and skidding.

Navigation results of three measurement models are shown in Figs. 13-18. Fig. 13 gives the estimated biases for gyroscopes and accelerometers of three methods in the first run. It indicates that the upward biases are more unstable and larger than those in other directions. Fig. 14 provides the estimation results of the scale factor and two mounting angles. Results are similar for three methods in two runs except that the mounting angle $\theta$ in the first run is more unstable, and two mounting angles were slightly different in two runs. In Fig. 15, the lever arm estimation results of three methods are inconsistent to each other; however, the system performance is not significantly affected by this inconsistency. We found that the lever arm can be regarded as a kind of 'error buffer' to account for the unconsidered uncertainties of the measurement errors. This observation is a good support for online calibration of the involved parameters first advocated by our group [13].

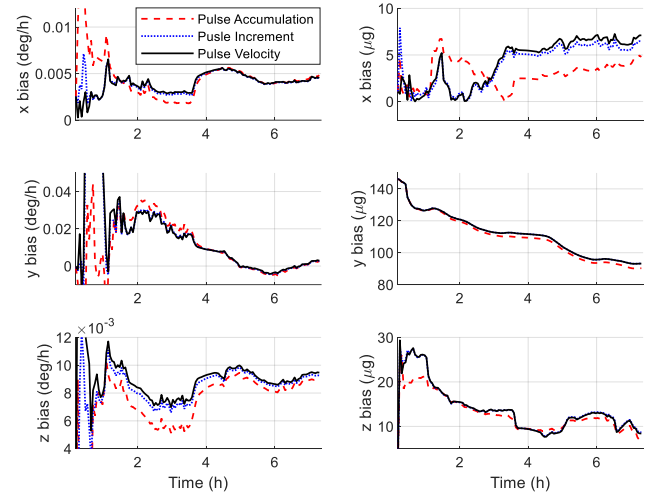

Figure 13: Estimation results of gyroscopes and accelerometers for the first run (Results of the second run are similar).

The position estimation errors are shown in Figs. 16-17. It can be seen that pulse velocity measurements yield the best accuracy than the other two kinds of measurements do. The relative horizontal position errors are mostly lower than 0.2‰ of the travelled distance. In contrast, the relative position errors of the pulse accumulation and increment measurements are approximately 0.2-0.4‰ of the travelled distance. As for the orientation estimation accuracy, three methods generate similar results in Fig. 18.

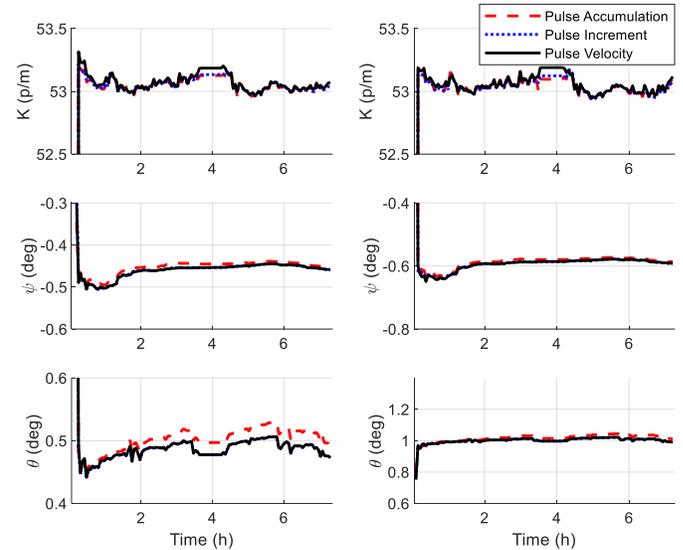

Figure 14: Estimation results of odometer scale factor and mounting angles. The left column is for the first run, and the right column is for the second run.

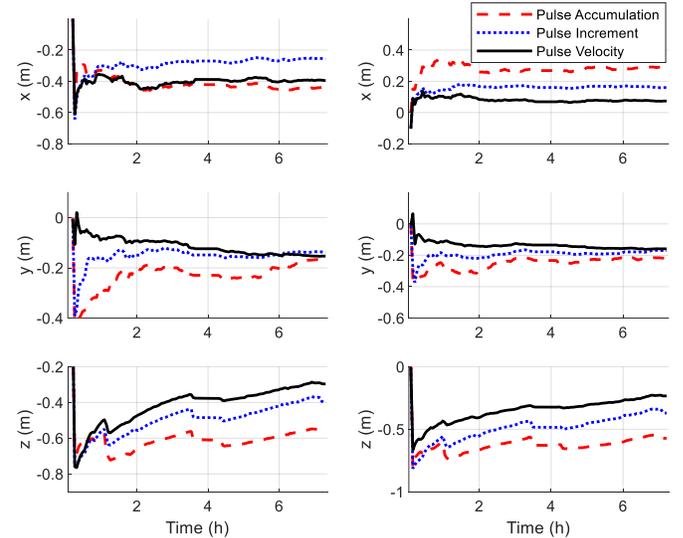

Figure 15: Estimation results of the lever arm. The left column is for the first run, and the right column is for the second run.

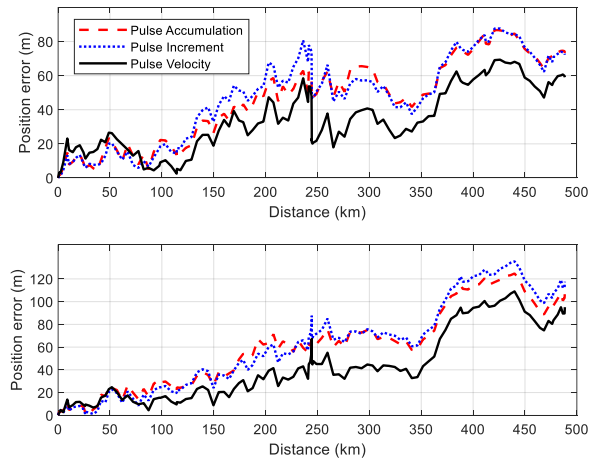

Figure 16: The absolute position estimation errors for three methods. The top is for the first run, and the bottom is for the second run.

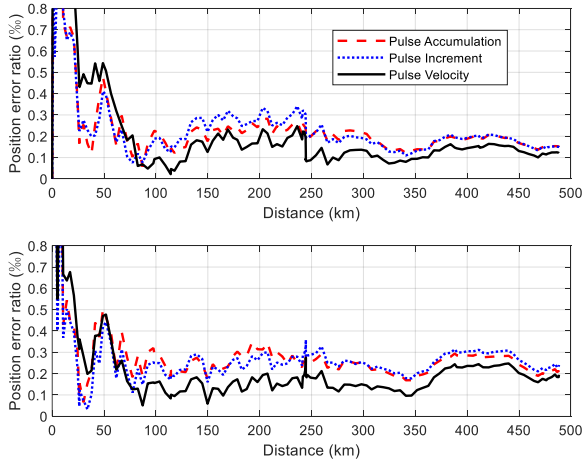

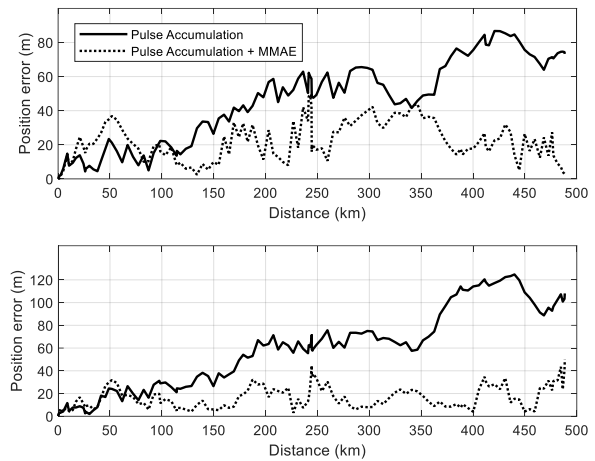

Figure 17: The relative position errors of three methods. The top is for the first run, and the bottom is for the second run.

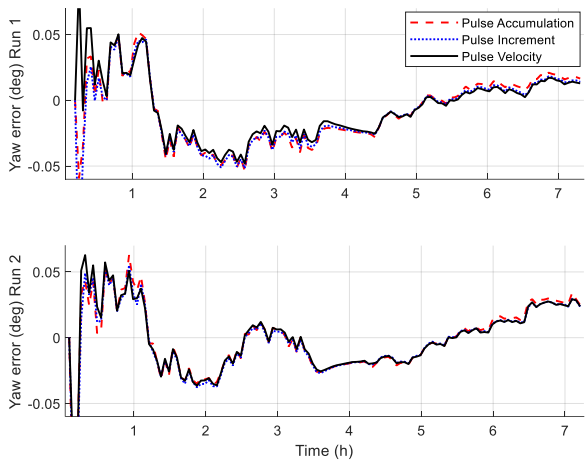

Figure 18: Yaw angle estimation errors of three methods for two runs.

Figure 19: The position estimation errors by MMAE on the pulse accumulation measurement. The top is for the first run, and the bottom is for the second run.

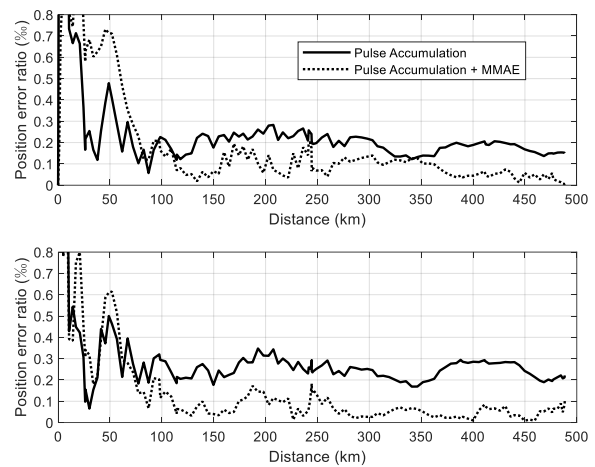

Figure 20: Relative errors by MMAE on the pulse accumulation measurement. The top is for the first run, and the bottom is for the second run.

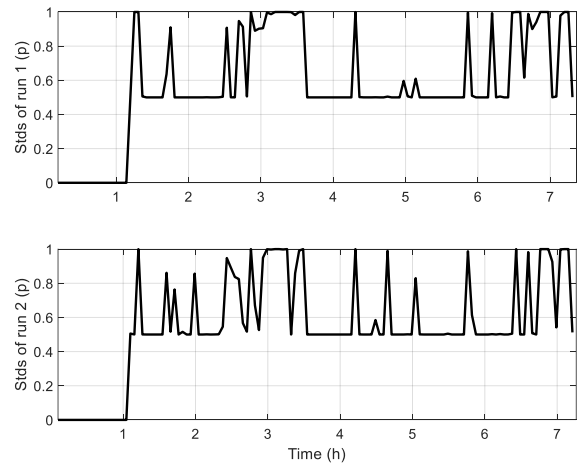

Figure 21: Adapted odometer pulse increment error stds for two runs.

### B. Results of MMAE

The difficulty of using the MMAE resides in designing the parameter models. In this work, the standard deviations (stds) of three kinds of the odometer measurement errors are respectively designed in the MMAE, and the theoretical analyses provided in Section II lead to convenient selection of possible stds. The interacting multiple model (IMM) estimation method [27], [36] was also tested but was inferior to the MMAE method for the current problem. It agrees with the recommendation in [37] that the MMAE is more stable in parameter adaptation, and thus more preferred in the selection of suitable error stds.

#### 1) MMAE on pulse accumulation measurement

According to the error statistics of the pulse accumulation measurements, the MMAE models here included stds {0.5, 0.6, 0.7, 0.8, 0.9, 1}. Position estimation results are compared in Figs. 19-20. It can be seen that the improvements of MMAE over the standard pulse accumulation method are significant, especially for the long-distance stability of position errors. Fig. 21 gives the adapted stds of MMAE in the two runs.

#### 2) MMAE on pulse increment measurement

The ideal error characteristics of the incremental pulse measurement have been given in Lemma 2. However, including larger measurement stds for incremental pulses was found to be more helpful in the experiments. In our tests, the bank of models was assigned with stds from the set {0.5, 1, 2}. Results of MMAE are shown in Figs. 22-23. It can be seen that horizontal position errors are obviously ameliorated. As shown in Fig. 24, the MMAE method frequently adjusts to select the most suitable stds. Most of the time, the optimal std is small, which means that the pulse increment measurement error is acceptable.

### 3) MMAE on pulse velocity measurement

Varying acceleration and deceleration are inevitable in practical driving. In addressing large measurement errors, the aforementioned variance relaxation method is quite ad hoc. In contrast, the MMAE approach provides a more delicate treatment of this problem. The bank of stds was designed as {1, 2, 3, 5}, in view of the above observation that high-accuracy pulse velocity information could be obtained most of the time. Results presented in Figs. 25-26 show the improvement on the horizontal position over the original EKF method. Fig. 27 gives the adapted stds, which indicates that the pulse velocity measurements are accurate in most cases.

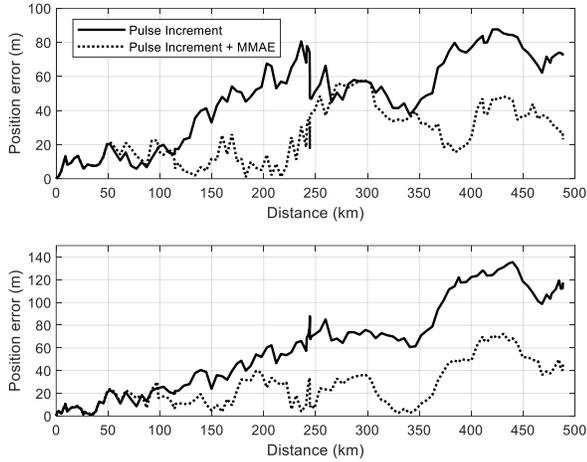

Figure 22: The position estimation errors by MMAE on the pulse increment measurement. The top is for the first run, and the bottom is for the second run.

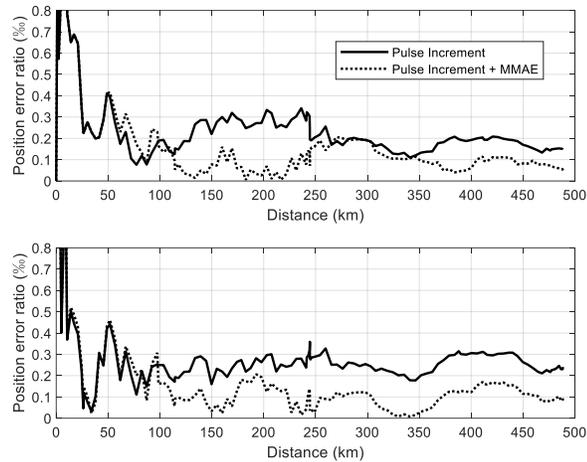

Figure 23: Relative errors by MMAE on the pulse increment measurement. The top is for the first run, and the bottom is for the second run.

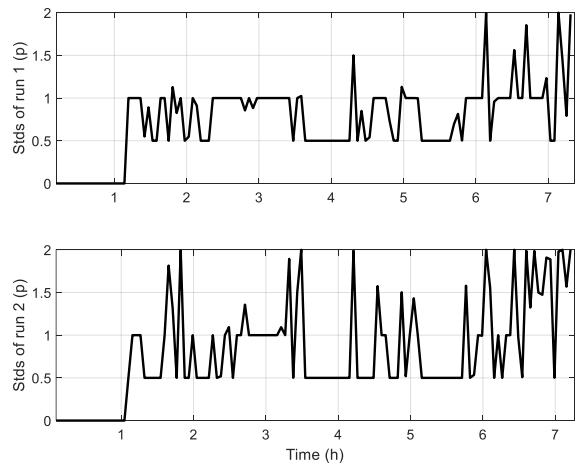

Figure 24: Adapted odometer pulse increments error stds for two runs.

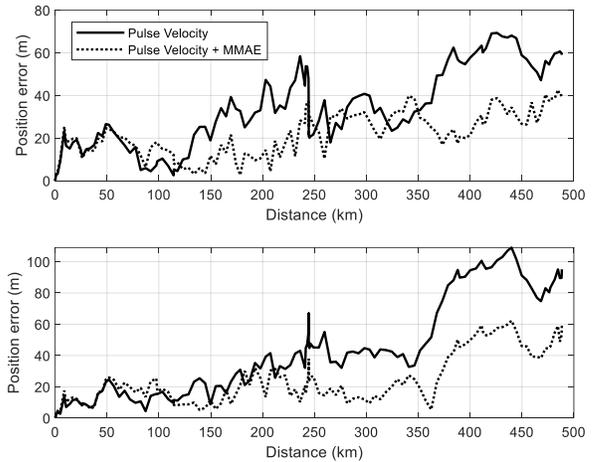

Figure 25: The position estimation errors by MMAE on the pulse velocity measurement. The top is for the first run, and the bottom is for the second run.

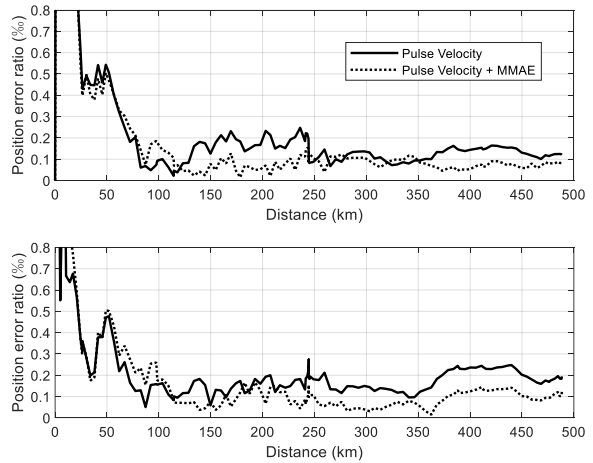

Figure 26: Relative errors by MMAE on the pulse velocity measurement. The top is for the first run, and the bottom is for the second run.

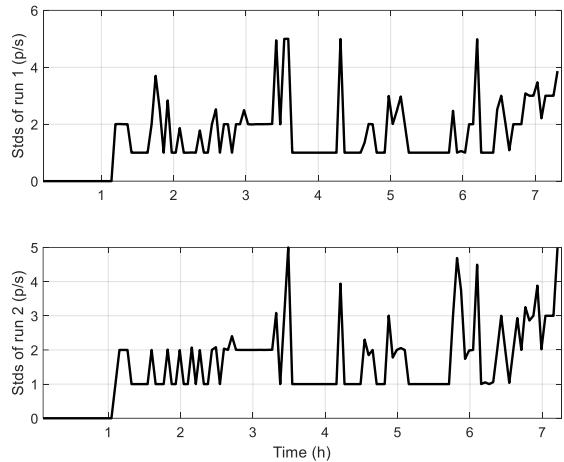

Figure 27: Adapted odometer pulse velocity error stds for two runs.

## C. Detailed Measures for Performance Evaluation

In Fig. 28, we also compared the position estimation results of three measurements after using the MMAE. It can be seen that the long-distance position accuracy of pulse accumulation measurement is slightly better than the others. In order to further quantitatively assess the performance of the proposed methods, three measures are presented here. The results with the measures defined below are given in Table II. Note that the absolute error is in meter and the relative error is the ratio of the position error w.r.t. the travelled distance.

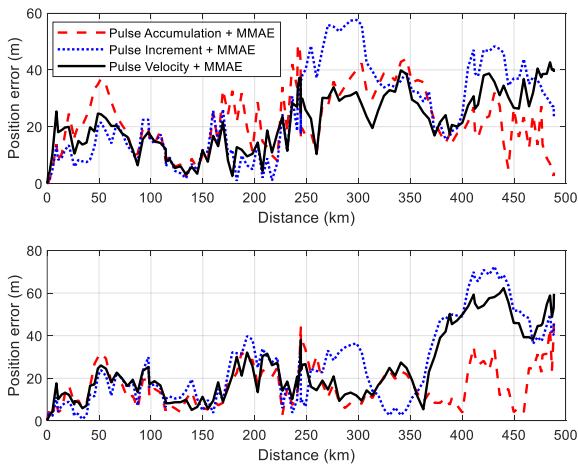

Figure 28: Comparison of position errors after using MMAE. The top is for the first run, and the bottom is for the second run.

### 1) Mean position error (Mean20)

We observed that the position error often becomes stable after 20km. The criterion *Mean20* is thus defined as the average of the position errors after 20km. The criterion reflects the overall performance, and the mean absolute error and the mean relative error are both given.

### 2) Gradient (Gra20)

A straight line is used to fit the position errors after 20km. The criterion *Gra20* is defined as the slope of the line. This criterion indicates the stability of errors. Specifically, both the absolute error and the relative error are computed. The smaller the slope, the better the estimation error stability.

### 3) Maximum (Max20)

This measure is defined as the maximum position error after 20km. Note that this measure is only computed with the absolute error, since the relative error ratio at 20km is always the largest.

TABLE II
PERFORMANCE MEASURES OF PROPOSED METHODS

| Test | Methods | Mean20 (m, ‰) | | Gra20 (m, ‰) | | Max20(m) |
|------|---------|-------|-------|---------|----------|----------|
| 1 | PA | 48.05 | 0.202 | 0.156 | -1.77e-4 | 88.67 |
| 2 | PA | 62.98 | 0.252 | 0.230 | -6.94e-5 | 128.7 |
| 1 | PI | 49.64 | 0.209 | 0.151 | -1.90e-4 | 90.69 |
| 2 | PI | 66.20 | 0.252 | 0.264 | 1.19 e-4 | 138.7 |
| 1 | PV | **34.55** | **0.168** | **0.112** | **-4.47e-4** | **72.58** |
| 2 | PV | **46.91** | **0.190** | **0.201** | **-8.27e-5** | **111.4** |
| 1 | PA_M | **22.03** | 0.154 | -6e-4 | **-9.25e-4** | 51.28 |
| 2 | PA_M | **19.86** | 0.119 | **0.019** | **-5.63e-4** | 49.84 |
| 1 | PI_M | 26.48 | **0.127** | 0.068 | -3.17e-4 | 59.86 |
| 2 | PI_M | 26.17 | **0.118** | 0.096 | -2.12e-4 | 72.47 |
| 1 | PV_M | 23.60 | 0.132 | 0.053 | -5.44e-4 | 54.57 |
| 2 | PV_M | 26.59 | 0.132 | 0.088 | -3.75e-4 | 69.23 |

PA: Pulse accumulation. PI: Pulse increment. PV: Pulse velocity. The suffix 'M' denotes the MMAE-aided versions. For Mean20 and Gra20, the left column was computed with absolute errors, and the right column was computed with relative errors.

Here we adopt the absolute mean and the stability to assess the algorithms. Among three types of measurements, Table II indicates that the performance of the standard pulse velocity measurement is about 39% better in mean, and 28% better in stability. But aided by MMAE, the pulse accumulation measurement slightly outperforms the other two types of measurements about 23% in mean, whereas it is more stable by over 7 times. This indicates that the contribution of MMAE is more significant to the pulse accumulation measurement than the other two measurements.

In summary, the properties of three measurements are finally concluded in Table III, which might be conducive to their practical applications. The error property is corresponding to the theoretical analyses in Section III. Considering more computations are required for the MMAE-aided EKF, we regard the standard EKF versions as 'fast' and name the MMAE-aided versions as 'effective'. Moreover, the pulse velocity measurement with standard EKF and the pulse accumulation measurement with MMAE-EKF are both denoted as 'recommended'.

TABLE III
SUMMARY OF THREE MEASUREMENTS

| Methods | Error Property | Standard EKF | MMAE-EKF |
|---------|----------------|--------------|----------|
| PA | Motion-dependent | fast | Effective & Recommended |
| PI | Motion-dependent & Correlated | fast | Effective |
| PV | Motion-dependent | fast & Recommended | Effective |

PA: Pulse accumulation. PI: Pulse increment. PV: Pulse velocity.

## VII. CONCLUSION

In this article, the INS and odometer integrated navigation algorithms were investigated, focusing on rigorous error analysis of the odometer pulse and the countermeasures.

Specifically, three types of measurements were formulated based on the pulse accumulation, pulse increment, and pulse-derived velocity, respectively. In addition, the multiple model estimation method was applied to further improve the performance by accounting for the motion-dependent measurement errors. Field tests were repeated on the same route about 490km for two times. The average position errors of three types of measurements are all better than 0.25‰ of travelled distance. Moreover, results of simulations and field tests show that the pulse velocity measurement performs the best in standard EKF realizations. After incorporating the MMAE approach, the average position errors of three types of measurements are all better than 0.15‰ of travelled distance, and the pulse accumulation measurement is more privileged in the sense of error stability, despite more computation-intensive.

The delicate manipulation of odometer pulses proposed in this paper is hopefully beneficial to other wheeled applications, such as the land robotics navigation and the pipeline survey, etc.

## Appendix A

The pulse velocity error in (27) is simplified by only considering the error of each state.

$$
\begin{aligned}
\delta \dot{s} = & -K \mathbf{e}_1^T \mathbf{C}_n^m \Big[ \mathbf{C}_n^b \big( \mathbf{v}^n \times \big) + \mathbf{l}^b \times \mathbf{C}_n^b \big( \boldsymbol{\omega}_{ie}^n \times \big) \Big] \boldsymbol{\phi}^n \\
& + K \mathbf{e}_1^T \mathbf{C}_b^m \mathbf{C}_n^b \delta \mathbf{v}_I^n - K \mathbf{e}_1^T \mathbf{C}_b^m \big( \mathbf{l}^b \times \big) \mathbf{b}_g \\
& + K \mathbf{e}_1^T \mathbf{C}_b^m \big( \mathbf{l}^b \times \big) \mathbf{C}_n^b \omega_{ie} \begin{bmatrix} 0 & -\sin L & 0 \\ 0 & \cos L & 0 \\ 0 & 0 & 0 \end{bmatrix} \delta \mathbf{p} \\
& + K \mathbf{e}_1^T \mathbf{M}_3 \big( \theta \big) \mathbf{D}_{M_2} \big( \psi \big) \big( \mathbf{C}_n^b \mathbf{v}_I^n + \boldsymbol{\omega}_{eb}^b \times \mathbf{l}^b \big) \delta \psi \\
& + K \mathbf{e}_1^T \mathbf{D}_{M_3} \big( \theta \big) \mathbf{M}_2 \big( \psi \big) \big( \mathbf{C}_n^b \mathbf{v}_I^n + \boldsymbol{\omega}_{eb}^b \times \mathbf{l}^b \big) \delta \theta \\
& + K \mathbf{e}_1^T \mathbf{C}_b^m \boldsymbol{\omega}_{eb}^b \times \delta \mathbf{l}^b + \mathbf{e}_1^T \mathbf{C}_b^m \big( \mathbf{C}_n^b \mathbf{v}_I^n + \boldsymbol{\omega}_{eb}^b \times \mathbf{l}^b \big) \delta K
\end{aligned}
\tag{49}
$$

in which, the mounting matrix is denoted as

$$
\mathbf{C}_b^m = \mathbf{M}_3 \big( \theta \big) \mathbf{M}_2 \big( \psi \big)
\tag{50}
$$

And, the derivatives of the elementary rotation matrix are given as

$$
\mathbf{D}_{M_2} (\psi) = \begin{bmatrix} -\sin \psi & 0 & -\cos \psi \\ 0 & 0 & 0 \\ \cos \psi & 0 & -\sin \psi \end{bmatrix}, \mathbf{D}_{M_3} (\theta) = \begin{bmatrix} -\sin \theta & \cos \theta & 0 \\ -\cos \theta & -\sin \theta & 0 \\ 0 & 0 & 0 \end{bmatrix}
\tag{51}
$$

Therefore, the partial derivatives can be computed as follows

$$
\mathbf{M}_k (1:3) = \frac{\partial \delta \dot{s}}{\partial \boldsymbol{\phi}^{nT}} = -K \mathbf{e}_1^T \mathbf{C}_b^m \Big[ \mathbf{C}_n^b \big( \mathbf{v}^n \times \big) + \mathbf{l}^b \times \mathbf{C}_n^b \big( \boldsymbol{\omega}_{ie}^n \times \big) \Big]
\tag{52}
$$

$$
\mathbf{M}_k (4:6) = \frac{\partial \delta \dot{s}}{\partial \mathbf{v}^{nT}} = K \mathbf{e}_1^T \mathbf{C}_b^m \mathbf{C}_n^b
\tag{53}
$$

$$
\mathbf{M}_k (7:9) = \frac{\partial \delta \dot{s}}{\partial \mathbf{p}^T} = K \mathbf{e}_1^T \mathbf{C}_b^m \big( \mathbf{l}^b \times \big) \mathbf{C}_n^b \omega_{ie} \begin{bmatrix} 0 & -\sin L & 0 \\ 0 & \cos L & 0 \\ 0 & 0 & 0 \end{bmatrix}
\tag{54}
$$

$$
\mathbf{M}_k (10:12) = \frac{\partial \delta \dot{s}}{\partial \mathbf{b}_g^T} = -K \mathbf{e}_1^T \mathbf{C}_b^m \big( \mathbf{l}^b \times \big)
\tag{55}
$$

$$
\mathbf{M}_k (16) = \frac{\partial \delta \dot{s}}{\partial \delta K} = \mathbf{e}_1^T \mathbf{C}_b^m \big( \mathbf{C}_n^b \mathbf{v}_I^n + \boldsymbol{\omega}_{eb}^b \times \mathbf{l}^b \big)
\tag{56}
$$

$$
\mathbf{M}_k (17) = \frac{\partial \delta \dot{s}}{\partial \delta \psi} = K \mathbf{e}_1^T \mathbf{M}_3 \big( \theta \big) \mathbf{D}_{M_2} \big( \psi \big) \big( \mathbf{C}_n^b \mathbf{v}_I^n + \boldsymbol{\omega}_{eb}^b \times \mathbf{l}^b \big)
\tag{57}
$$

$$
\mathbf{M}_k (18) = \frac{\partial \delta \dot{s}}{\partial \delta \theta} = K \mathbf{e}_1^T \mathbf{D}_{M_3} \big( \theta \big) \mathbf{M}_2 \big( \psi \big) \big( \mathbf{C}_n^b \mathbf{v}_I^n + \boldsymbol{\omega}_{eb}^b \times \mathbf{l}^b \big)
\tag{58}
$$

$$
\mathbf{M}_k (19:21) = \frac{\partial \delta \dot{s}}{\partial \delta \mathbf{l}^{bT}} = K \mathbf{e}_1^T \mathbf{C}_b^m \big( \boldsymbol{\omega}_{eb}^b \times \big)
\tag{59}
$$

Finally, the Jacobian matrix of the pulse velocity model w.r.t. the error state is

$$
\mathbf{F}_p = \big[ \mathbf{M}_k, 0 \big]^T
\tag{60}
$$

## Appendix B

This appendix gives detailed derivation of the velocity integration in (31). According to the velocity integration method proposed in [38], the terms related to attitude are rewritten as

$$
\mathbf{C}_n^b (t) \mathbf{v}^n (t) = \mathbf{C}_{b(t_k)}^{b(t)} \mathbf{C}_{n(t_k)}^{b(t_k)} \mathbf{C}_{n(t)}^{n(t_k)} \mathbf{v}^n (t)
\tag{61}
$$

$$
\begin{aligned}
\big( \mathbf{C}_n^b (t) \boldsymbol{\omega}_{ie}^n \big) \times \mathbf{l}^b & = \mathbf{C}_n^b (t) \big( \boldsymbol{\omega}_{ie}^n \times \mathbf{C}_b^n (t) \mathbf{l}^b \big) \\
& = \mathbf{C}_{b(t_k)}^{b(t)} \mathbf{C}_{n(t_k)}^{b(t_k)} \mathbf{C}_{n(t)}^{n(t_k)} \big( \boldsymbol{\omega}_{ie}^n \times \big) \mathbf{C}_{n(t_k)}^{n(t)} \mathbf{C}_{b(t_k)}^{n(t_k)} \mathbf{C}_{b(t)}^{b(t_k)} \mathbf{l}^b
\end{aligned}
\tag{62}
$$

And, we have

$$
\begin{aligned}
\mathbf{C}_{b(t)}^{b(t_k)} & = \mathbf{I} - \Big( \int_{t_k}^t \boldsymbol{\omega}_{ib}^b dt \Big) \times \\
& \approx \mathbf{I} - \Big( \frac{(t - t_k)^2}{2} \mathbf{a}_\omega + (t - t_k) \mathbf{b}_\omega \Big) \times
\end{aligned}
\tag{63}
$$

$$
\mathbf{C}_{n(t)}^{n(t_k)} \approx \mathbf{I} + (t - t_k) \boldsymbol{\omega}_{in}^n \times
\tag{64}
$$

where $\mathbf{a}_\omega, \mathbf{b}_\omega$ are computed with two samples of gyroscopes.

$$
\mathbf{a}_\omega = \frac{4 \big( \Delta \boldsymbol{\theta}_2 - \Delta \boldsymbol{\theta}_1 \big)}{T^2}
\tag{65}
$$

$$
\mathbf{b}_\omega = \frac{3 \Delta \boldsymbol{\theta}_1 - \Delta \boldsymbol{\theta}_2}{T}
$$

Substitute (63), (64) into (61), the integration of (61) is computed as

$$
\begin{aligned}
& \int_{t_k}^{t_{k+1}} \mathbf{C}_{b(t_k)}^{b(t)} \mathbf{C}_{n(t_k)}^{b(t_k)} \mathbf{C}_{n(t)}^{n(t_k)} \mathbf{v}^n (t) dt \\
& = \int_{t_k}^{t_{k+1}} \mathbf{C}_{n(t_k)}^{b(t_k)} \mathbf{v}^n (t) dt + \int_{t_k}^{t_{k+1}} (t - t_k) \mathbf{C}_{n(t_k)}^{b(t_k)} \boldsymbol{\omega}_{in}^n \times \mathbf{v}^n (t) dt \\
& - \int_{t_k}^{t_{k+1}} \Big( \frac{(t - t_k)^2}{2} \mathbf{a}_\omega + (t - t_k) \mathbf{b}_\omega \Big) \times \mathbf{C}_{n(t_k)}^{b(t_k)} \mathbf{v}^n (t) dt \\
& - \int_{t_k}^{t_{k+1}} \Big( \frac{(t - t_k)^3}{2} \mathbf{a}_\omega + (t - t_k)^2 \mathbf{b}_\omega \Big) \times \mathbf{C}_{n(t_k)}^{b(t_k)} \boldsymbol{\omega}_{in}^n \times \mathbf{v}^n (t) dt
\end{aligned}
\tag{66}
$$

Suppose the velocity in $[t_k, t_{k+1}]$ changes linearly, i.e.,

$$\mathbf{v}^n(t) = \mathbf{v}^n(t_k) + \frac{t - t_k}{T}\left(\mathbf{v}^n(t_{k+1}) - \mathbf{v}^n(t_k)\right) \quad (67)$$

The integrations in (66) are further approximated as

$$
\begin{aligned}
&\int_{t_k}^{t_{k+1}} \mathbf{C}_{b(t_k)}^{b(t)} \mathbf{C}_{n(t_k)}^{b(t_k)} \mathbf{C}_{n(t)}^{n(t_k)} \mathbf{v}^n(t)\,dt \\
&\approx \frac{T}{2}\mathbf{C}_{n(t_k)}^{b(t_k)}\left(\mathbf{v}^n(t_{k+1}) + \mathbf{v}^n(t_k)\right) + \frac{T^2}{6}\mathbf{C}_{n(t_k)}^{b(t_k)}\boldsymbol{\omega}_{in}^n \times \mathbf{v}^n(t_k) \\
&+ \frac{T^2}{3}\mathbf{C}_{n(t_k)}^{b(t_k)}\boldsymbol{\omega}_{in}^n \times \mathbf{v}^n(t_{k+1}) - \frac{T}{3}\Delta\boldsymbol{\theta}_1 \times \mathbf{C}_{n(t_k)}^{b(t_k)}\mathbf{v}^n(t_k) \\
&- \frac{T}{6}\left(3\Delta\boldsymbol{\theta}_1 + \Delta\boldsymbol{\theta}_2\right) \times \mathbf{C}_{n(t_k)}^{b(t_k)}\mathbf{v}^n(t_{k+1}) \\
&- \frac{T^2}{60}\left(9\Delta\boldsymbol{\theta}_1 - \Delta\boldsymbol{\theta}_2\right) \times \mathbf{C}_{n(t_k)}^{b(t_k)}\boldsymbol{\omega}_{in}^n \times \mathbf{v}^n(t_k) \\
&- \frac{T^2}{20}\left(7\Delta\boldsymbol{\theta}_1 + 3\Delta\boldsymbol{\theta}_2\right) \times \mathbf{C}_{n(t_k)}^{b(t_k)}\boldsymbol{\omega}_{in}^n \times \mathbf{v}^n(t_{k+1})
\end{aligned}
\quad (68)
$$

Similarly, the integration of (62) is approximated as

$$
\begin{aligned}
&\int_{t_k}^{t_{k+1}}\left(\mathbf{C}_{b(t_k)}^{b(t_k)}\mathbf{C}_{n(t_k)}^{b(t_k)}\mathbf{C}_{n(t)}^{n(t_k)}\boldsymbol{\omega}_{ie}^n \times \mathbf{C}_{n(t_k)}^{n(t)}\mathbf{C}_{b(t_k)}^{n(t_k)}\mathbf{C}_{b(t)}^{b(t_k)}\right)\mathbf{l}^b\,dt \\
&= \int_{t_k}^{t_{k+1}}\left(\mathbf{I} - \left[\frac{(t-t_k)^3}{2}\mathbf{a}_\omega + (t-t_k)^2\,\mathbf{b}_\omega\right]\times\right)\mathbf{C}_{n(t_k)}^{b(t_k)} \\
&\times \left(\mathbf{I} + (t - t_k)\boldsymbol{\omega}_{in}^n \times\right)\cdot\left(\boldsymbol{\omega}_{ie}^n\times\right)\cdot\left(\mathbf{I} - (t - t_k)\boldsymbol{\omega}_{in}^n\times\right)\mathbf{C}_{b(t_k)}^{n(t_k)} \\
&\times \left(\mathbf{I} + \left(\frac{(t-t_k)^3}{2}\mathbf{a}_\omega + (t-t_k)^2\,\mathbf{b}_\omega\right)\times\right)\mathbf{l}^b\,dt
\end{aligned}
\quad (69)
$$

In view of the fact that $\boldsymbol{\omega}_{ie}^n$ and $\boldsymbol{\omega}_{in}^n$ are both in the order of $10^{-5}$. Therefore, their multiplication can be omitted, and (69) is approximated as

$$
\begin{aligned}
&\int_{t_k}^{t_{k+1}}\left(\mathbf{C}_{b(t_k)}^{b(t_k)}\mathbf{C}_{n(t_k)}^{b(t_k)}\mathbf{C}_{n(t)}^{n(t_k)}\left(\boldsymbol{\omega}_{ie}^n\times\right)\mathbf{C}_{n(t_k)}^{n(t)}\mathbf{C}_{b(t_k)}^{n(t_k)}\mathbf{C}_{b(t)}^{b(t_k)}\right)\mathbf{l}^b\,dt \\
&\approx T\cdot\mathbf{C}_{n(t_k)}^{b(t_k)}\boldsymbol{\omega}_{ie}^n\times\mathbf{C}_{b(t_k)}^{n(t_k)}\mathbf{l}^b + \frac{T^2}{2}\mathbf{C}_{n(t_k)}^{b(t_k)}\boldsymbol{\omega}_{ie}^n\times\mathbf{C}_{b(t_k)}^{n(t_k)}\boldsymbol{\omega}_{ib}^b\times\mathbf{l}^b \\
&- \frac{T}{6}\left(5\Delta\boldsymbol{\theta}_1 + \Delta\boldsymbol{\theta}_2\right)\times\mathbf{C}_{n(t_k)}^{b(t_k)}\boldsymbol{\omega}_{ie}^n\times\mathbf{C}_{b(t_k)}^{n(t_k)}\mathbf{l}^b \\
&- \frac{T^3}{40}\left(7\Delta\boldsymbol{\theta}_1 + 3\Delta\boldsymbol{\theta}_2\right)\times\mathbf{C}_{n(t_k)}^{b(t_k)}\boldsymbol{\omega}_{ie}^n\times\mathbf{C}_{b(t_k)}^{n(t_k)}\mathbf{a}_\omega\times\mathbf{l}^b \\
&- \frac{T^2}{6}\left(3\Delta\boldsymbol{\theta}_1 + \Delta\boldsymbol{\theta}_2\right)\times\mathbf{C}_{n(t_k)}^{b(t_k)}\boldsymbol{\omega}_{ie}^n\times\mathbf{C}_{b(t_k)}^{n(t_k)}\mathbf{b}_\omega\times\mathbf{l}^b
\end{aligned}
\quad (70)
$$

The integration of the middle term in (31) is approximated as

$$\int_{t_k}^{t_{k+1}}\boldsymbol{\omega}_{ib}^b(t)\times\mathbf{l}^b\,dt \approx \left(\Delta\boldsymbol{\theta}_1 + \Delta\boldsymbol{\theta}_2\right)\times\mathbf{l}^b \quad (71)$$

Since the integration interval $T$=0.02s or more smaller, small terms are further omitted and the integration of (31) are finally approximated as

$$
\begin{aligned}
&\int_{t_k}^{t_{k+1}}\mathbf{C}_n^b(t)\mathbf{v}^n(t) + \boldsymbol{\omega}_{eb}^b(t)\times\mathbf{l}^b\,dt \\
&\approx \frac{T}{2}\mathbf{C}_{n(t_k)}^{b(t_k)}\left(\mathbf{v}^n(t_{k+1}) + \mathbf{v}^n(t_k)\right) + \left(\Delta\boldsymbol{\theta}_1 + \Delta\boldsymbol{\theta}_2\right)\times\mathbf{l}^b \\
&- \frac{T}{3}\Delta\boldsymbol{\theta}_1\times\mathbf{C}_{n(t_k)}^{b(t_k)}\mathbf{v}^n(t_k) - \frac{T}{6}\left(3\Delta\boldsymbol{\theta}_1 + \Delta\boldsymbol{\theta}_2\right)\times\mathbf{C}_{n(t_k)}^{b(t_k)}\mathbf{v}^n(t_{k+1}) \\
&- T\mathbf{C}_{n(t_k)}^{b(t_k)}\boldsymbol{\omega}_{ie}^n\times\mathbf{C}_{b(t_k)}^{n(t_k)}\mathbf{l}^b - \frac{T}{6}\left(5\Delta\boldsymbol{\theta}_1 + \Delta\boldsymbol{\theta}_2\right)\times\mathbf{C}_{n(t_k)}^{b(t_k)}\boldsymbol{\omega}_{ie}^n\times\mathbf{C}_{b(t_k)}^{n(t_k)}\mathbf{l}^b
\end{aligned}
\quad (72)
$$

## Appendix C

The Jacobian matrix $\mathbf{M}_k$ has been given in (60). Therefore, $\mathbf{H}_k$ in (35) is derived by integrating the elements of $\mathbf{M}_k$ as

$$
\begin{aligned}
M_\phi &= \int_{t_k}^{t_{k+1}}\frac{\partial\delta\dot{s}}{\partial\phi^{nT}}dt \\
&= -K\mathbf{e}_1^T\mathbf{C}_b^m\left[\int_{t_k}^{t_{k+1}}\mathbf{C}_n^b\left(\mathbf{v}^n\times\right)dt + \mathbf{l}^b\times\int_{t_k}^{t_{k+1}}\mathbf{C}_n^b\,dt\left(\boldsymbol{\omega}_{ie}^n\times\right)\right]
\end{aligned}
\quad (73)
$$

$$M_{\mathbf{v}} = \int_{t_k}^{t_{k+1}}\frac{\partial\delta\dot{s}}{\partial\delta\mathbf{v}^{nT}}dt = K\mathbf{e}_1^T\mathbf{C}_b^m\int_{t_k}^{t_{k+1}}\mathbf{C}_n^b\,dt \quad (74)$$

$$
\begin{aligned}
M_{\mathbf{p}} &= \int_{t_k}^{t_{k+1}}\frac{\partial\delta\dot{s}}{\partial\delta\mathbf{p}^{T}}dt \\
&= K\mathbf{e}_1^T\mathbf{C}_b^m\left(\mathbf{l}^b\times\right)\int_{t_k}^{t_{k+1}}\mathbf{C}_n^b\,dt\,\omega_{ie}\begin{bmatrix} 0 & -\sin L & 0 \\ 0 & \cos L & 0 \\ 0 & 0 & 0 \end{bmatrix}
\end{aligned}
\quad (75)
$$

$$M_{\mathbf{b}} = \int_{t_k}^{t_{k+1}}\frac{\partial\delta\dot{s}}{\partial\delta\mathbf{b}_g^{T}}dt = -K\mathbf{e}_1^T\mathbf{C}_b^m\left(\mathbf{l}^b\times\right)T \quad (76)$$

$$M_K = \int_{t_k}^{t_{k+1}}\frac{\partial\delta\dot{s}}{\partial\delta K}dt = \mathbf{e}_1^T\mathbf{C}_b^m\int_{t_k}^{t_{k+1}}\mathbf{C}_n^b\mathbf{v}^n + \boldsymbol{\omega}_{eb}^b\times\mathbf{l}^b\,dt \quad (77)$$

$$M_\psi = \int_{t_k}^{t_{k+1}}\frac{\partial\delta\dot{s}}{\partial\delta\psi}dt = K\mathbf{e}_1^T\mathbf{M}_3(\theta)\mathbf{D}_{M_2}(\psi)\int_{t_k}^{t_{k+1}}\mathbf{C}_n^b\mathbf{v}^n + \boldsymbol{\omega}_{eb}^b\times\mathbf{l}^b\,dt \quad (78)$$

$$M_\theta = \int_{t_k}^{t_{k+1}}\frac{\partial\delta\dot{s}}{\partial\delta\theta}dt = K\mathbf{e}_1^T\mathbf{D}_{M_3}(\theta)\mathbf{M}_2(\psi)\int_{t_k}^{t_{k+1}}\mathbf{C}_n^b\mathbf{v}^n + \boldsymbol{\omega}_{eb}^b\times\mathbf{l}^b\,dt \quad (79)$$

$$M_\mathbf{l} = \int_{t_k}^{t_{k+1}}\frac{\partial\delta\dot{s}}{\partial\delta\mathbf{l}^{bT}}dt = K\mathbf{e}_1^T\mathbf{C}_b^m\left(\boldsymbol{\omega}_{eb}^b\times\right)T \quad (80)$$

New integrations involved in computing the above elements are computed as

$$
\begin{aligned}
&\int_{t_k}^{t_k+T}\mathbf{C}_n^b\left(\mathbf{v}^n\times\right)dt = \int_{t_k}^{t_k+T}\mathbf{C}_{b(t_k)}^{b(t_k)}\mathbf{C}_{n(t_k)}^{b(t_k)}\mathbf{C}_{n(t)}^{n(t_k)}\left(\mathbf{v}^n\times\right)dt \\
&\approx \frac{T}{2}\mathbf{C}_{n(t_k)}^{b(t_k)}\left(\mathbf{v}^n(t_{k+1}) + \mathbf{v}^n(t_k)\right)\times \\
&+ \frac{T}{3}\Delta\boldsymbol{\theta}_1\times\mathbf{C}_{n(t_k)}^{b(t_k)}\left(\mathbf{v}^n(t_k)\times\right) \\
&+ \frac{T}{6}\left(3\Delta\boldsymbol{\theta}_1 + \Delta\boldsymbol{\theta}_2\right)\times\mathbf{C}_{n(t_k)}^{b(t_k)}\left(\mathbf{v}^n(t_{k+1})\times\right)
\end{aligned}
\quad (81)
$$

$$\int_{t_k}^{t_{k+1}} \mathbf{C}_n^b dt = \int_{t_k}^{t_{k+1}} \mathbf{C}_{b(t_k)}^{b(t)} \mathbf{C}_{n(t_k)}^{b(t_k)} \mathbf{C}_{n(t)}^{n(t_k)} dt$$

$$\approx T \mathbf{C}_{n(t_k)}^{b(t_k)} - \frac{T}{6} \left( 5\Delta\boldsymbol{\theta}_1 + \Delta\boldsymbol{\theta}_2 \right) \times \mathbf{C}_{n(t_k)}^{b(t_k)}$$

(82)

Therefore,

$$\mathbf{H}_k = \begin{bmatrix} M_\phi, M_v, M_p, M_b, \mathbf{0}_{1\times3}, M_K, M_\psi, M_\theta, M_1 \end{bmatrix}^T$$

(83)